\documentclass[letterpaper]{article} 
\usepackage{mathtools}
\usepackage{aaai2026}  
\usepackage{times}  
\usepackage{helvet}  
\usepackage{courier}  
\usepackage[hyphens]{url}  
\usepackage{float}

\usepackage{graphicx} 
\usepackage{natbib}  
\usepackage{caption} 
\usepackage[ruled,vlined]{algorithm2e}
\usepackage{amssymb}
\usepackage{multirow}
\usepackage{enumitem}
\newcommand{\E}{\mathbb{E}}
\newcommand{\mE}{\mathbb{E}}
\newcommand{\Dcal}{\mathcal{D}}

\definecolor{MPLBlue}{HTML}{1F77B4}   
\definecolor{MPLGreen}{HTML}{019e73}  
\definecolor{MPLOrange}{HTML}{d55e00} 
\definecolor{MPLPurple}{HTML}{cb79a7} 

\usepackage{algorithmicx}

\usepackage{listings}
\DeclareCaptionStyle{ruled}{labelfont=normalfont,labelsep=colon,strut=off} 
\title{Stabilizing Policy Gradient Methods via Reward Profiling}

\author{
Shihab Ahmed\textsuperscript{\rm 1},
El Houcine Bergou\textsuperscript{\rm 4},
Yue Wang\textsuperscript{\rm 1,2},
Aritra Dutta\textsuperscript{\rm 2,3}
}

\affiliations{
\textsuperscript{\rm 1}Department of Electrical and Computer Engineering, University of Central Florida, FL, USA\\
\textsuperscript{\rm 2}Department of Computer Science, University of Central Florida, FL, USA\\
\textsuperscript{\rm 3}Department of Mathematics, University of Central Florida, FL, USA\\
\textsuperscript{\rm 4}College of Computing, Mohammed VI Polytechnic University, Ben Guerir, Morocco\\
\{Shihab.Ahmed, Yue.Wang, Aritra.Dutta\}@ucf.edu,\; elhoucine.bergou@um6p.ma
}

\usepackage{amsmath}

\usepackage{subcaption}
\usepackage{textcomp}

\usepackage{adjustbox}
\usepackage{amsthm}

\newtheorem{remark}{Remark}

\newtheorem{assumption}{Assumption}\newtheorem{theorem}{Theorem}\newtheorem{lemma}{Lemma}

\usepackage{soul}

\newcommand{\myNum}[1]{(\emph{#1})}
\newcommand{\smartparagraph}[1]{\vspace{2pt} \noindent {\bf #1}}

\usepackage{graphicx}

\usepackage{url}            
\usepackage{booktabs}       
\usepackage{amsfonts}       
\usepackage{nicefrac}       
\usepackage{microtype}      

\newcommand{\Mcal}{\mathcal{M}}
\newcommand{\Scal}{\mathcal{S}}
\newcommand{\Acal}{\mathcal{A}}

\begin{document}

\maketitle

\begin{abstract}
Policy gradient methods, which have been extensively studied in the last decade, offer an effective and efficient framework for reinforcement learning problems. However, their performances can often be unsatisfactory, suffering from unreliable reward improvements and slow convergence, due to high variance in gradient estimations. In this paper, we propose a universal reward profiling framework that can be seamlessly integrated with any policy gradient algorithm, where we selectively update the policy based on \emph{high-confidence performance estimations}. We theoretically justify that our technique will not slow down the convergence of the baseline policy gradient methods, but with high probability, will result in stable and monotonic improvements of their performance. Empirically, on eight continuous‐control benchmarks (Box2D and MuJoCo/PyBullet), our profiling yields up to $1.5\times$ faster convergence to near‐optimal returns, up to $1.75\times$ reduction in return variance on some setups. Our profiling approach offers a general, theoretically grounded path to more reliable and efficient policy learning in complex environments.
\end{abstract}

\begin{links}
    \link{Code}{https://github.com/shlhab/reward-profiling}
\end{links}

\section{Introduction}

Reinforcement learning (RL) optimizes an agent's performance in a stochastic, sequential decision-making task. Policy‐gradient (PG) methods, which directly optimize parameterized policies from sampled trajectories rather than relying on value‐function bootstrapping, form one of the core paradigms in RL  \citep{sutton1998reinforcement, schulman2015trust}. Its direct formulation hence enables PG's effective implementations in high‐dimensional and continuous‐control settings, including robotic manipulation \citep{peters2006policy}, locomotion \citep{todorov2012mujoco}, and autonomous driving \citep{schulman2017proximal}, where value‐based approaches are inefficient.

However, PG methods generally suffer from  \emph{high variance} in gradient estimations. Small stochastic fluctuations in early‐episode rewards of the trajectories can propagate through the Monte Carlo estimator (e.g.\ REINFORCE \citep{williams1992simple}), leading to erratic updates, slow convergence, and occasional collapse of performance \citep{ilyas2018closer, lehmann2024definitiveguidepolicygradients}.  Classic variance‐reduction techniques, including baseline/advantage normalization and trust‐region constraints, can improve the stability but also bring drawbacks: problem‐specific tuning \citep{chung2021beyond}, second‐order solvers \citep{schulman2015trust}, or extra computational overhead \citep{wu2017scalable}. A unified framework combining sample efficiency, stability, and simplicity that does not rely on problem-specific baselines can be useful. Therefore, a natural question arises: 
\begin{center}
{\textit{Is it possible to reduce PG variance and stabilize learning across arbitrary policy‐gradient methods, without relying on specialized tuning or heavy second‐order machinery?}}
\end{center}
In this paper, we provide an affirmative answer to this question by proposing our reward profiling framework. Our framework is based on a straightforward, natural idea: we only update the policy if the updated one implies a better performance. Under ideal circumstances, when our justifications are exact, our reward profiling will imply a monotonically increasing performance, addressing the issue of unstable performance in vanilla PG methods. 
\begin{figure}[t]
    \centering
    \includegraphics[width=0.84\columnwidth]{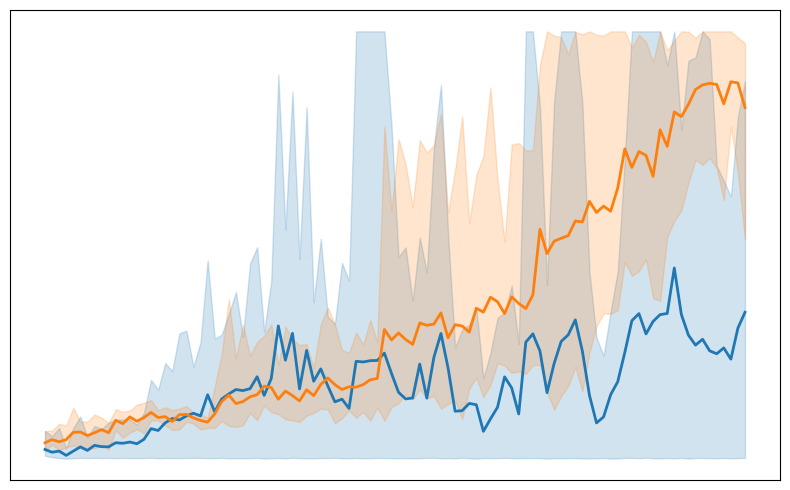}
    \caption{Training performance of REINFORCE. The agent converges with stable behavior with our reward profiling. 
    \textcolor{blue}{\textbf{—}} REINFORCE \quad
    \textcolor{orange}{\textbf{—}} REINFORCE+Lookback.}
    \label{fig:cartpole}
    \vspace{-5mm} 
\end{figure}
We perform a simple experiment to entertain this idea. We apply our reward profiling to a simple REINFORCE algorithm, and implement two algorithms under the CartPole environment \citep{brockman2016openai}. Figure~\ref{fig:cartpole} shows vanilla REINFORCE can suffer from unstable learning and severe performance fluctuations, whereas the reward profiling provides a much more stable, nearly monotonic improvement. This observation shows that the profiling framework can address the fundamental challenge in PG without acquiring any complicated techniques. 

In this paper, we further investigate this idea and develop our reward profiling framework to overcome the common instability issues of PG methods. Our contributions are summarized as follows.

\smartparagraph{Design of a universal reward-profiling framework (\S\ref{sec:framework}).} We first develop our universal reward-profiling framework in Algorithm \ref{alg:profiling-variants}, where we selectively accept, reject, or blend the policy updates based on the high‐confidence value estimations of their value functions. The framework requires a small number of additional rollouts per iteration, without incurring significant additional computation. First, we introduce our \texttt{Lookback} technique, which updates the policy only when the estimated cumulative reward of the new policy surpasses that of the current policy. We further develop two variants of our profiling technique (\texttt{Mix-up} and \texttt{Three-Points}) to address the potential issue of being stuck at a local optimum. They are designed to accelerate convergence while maintaining stability. This can be applied to any PG methods, and is expected to stabilize the learning process and improve their performance.

\smartparagraph{Theoretical guarantees (\S\ref{sec:convex case}).} 
We establish theoretical guarantees on the convergence and global optimality of the presented framework. We showed that, in the justification phase, with additional samples of order \(\mathcal{O}(\epsilon^{-2}\ln(T/\delta))\), we can compare the performances of the current and updated policies with high probability, thus ensuring monotonic improvements. We also show that the profiling framework will not slow down convergence theoretically, and enjoys an $\mathcal{O}(T^{-1/4})$ sub-optimality gap on the last iterate, providing a solid foundation for our proposed framework. 

\smartparagraph{Extensive empirical evaluation (\S\ref{sec:experiments}).} We adapt our framework with three representative algorithms: DDPG \citep{lillicrap2016continuous}, TRPO \citep{schulman2015trust}, and PPO \citep{schulman2017proximal}, and design corresponding profiling algorithms. We then evaluate their performance on continuous-control benchmarks. Results show consistent gains in final performance, sample efficiency, and training stability, providing validation for the algorithmic framework. We also port the algorithm to a Unity-ML “Reacher” DDPG agent with a separate simulation backend to verify that our framework is broadly applicable while yielding faster convergence and lower variance. 

\section{Related Work}

There are two lines of research aiming to tame high variance and improve stability in PG methods: \myNum{i} \emph{algorithm-centric} variants that bake variance-reduction or trust-region ideas into the core update, and \myNum{ii} \emph{wrapper-style} frameworks that layer on generic stability checks without re-engineering the underlying optimizer. 

 \myNum{i} \smartparagraph{Algorithm‐centric approaches.}
Early variance reduction in REINFORCE introduced \emph{baselines}, which are often a learned value function, to center Monte-Carlo returns, yielding unbiased and lower-variance updates \citep{sutton1999policy}.  Actor–critic architectures extend this idea by bootstrapping via temporal-difference learning, trading bias for further variance reduction \citep{konda1999actor, sutton1984temporal}; however, a misspecified critic step can introduce harmful bias \citep{chen2021closing, olshevsky2023smallgainanalysissingle}. As an alternative approach, trust-region methods such as TRPO enforce a KL-constraint to guarantee monotonic policy improvement under certain regularity conditions \citep{schulman2015trust}, but rely on second-order solvers, which are expensive.  PPO replaces TRPO’s conjugate-gradient and Hessian computations with a clipped surrogate objective, retaining many of the stability benefits while using only first-order updates \citep{schulman2017proximal}; nevertheless, it can still exhibit overly conservative steps or exploration failures in complex, contact-rich tasks \citep{wang2019trust}. Beyond on-policy methods, off-policy actor-critic algorithms like DDPG \citep{lillicrap2016continuous} and TD3 \citep{fujimoto18addressing} aim for high sample efficiency via replay buffers, but suffer from spiky Q-value estimates and divergent updates without careful regularization \citep{haarnoja2018soft, islam2017reproducibility}.  Variants such as SAC inject entropy bonuses for exploration and stability \citep{haarnoja2018soft}, yet their gains come at the cost of extra hyperparameters and temperature tuning.

\myNum{ii}\smartparagraph{Framework-based approaches.}
An orthogonal strand of work wraps a generic PG optimizer with lightweight checks or corrections. SVRG-style control variates (e.g.,\ SVRPG) freeze a reference policy to reduce gradient variance, at the expense of extra memory and on-policy rollouts \citep{xu2020improved}.~Momentum injections, such as Nesterov and heavy-ball variants, have been proposed to accelerate PG under smoothness assumptions \citep{xiao2022convergence, chen2024accelerated}; although they sometimes amplify early spikes.~In acceptance-based schemes, updates are only applied if a held-out performance estimate improves—trace back to optimistic or hysteretic updates in multi-agent settings \citep{omidshafiei2017deep, palma2018multiagent}; however, repeatedly validating a “true” return can be prohibitively costly.


Our reward-profiling wrapper unifies and extends these ideas with three simple, hyperparameter-minimal schemes:{\textbf{Lookback}} acts like a backtracking line search in policy space, rejecting any update whose empirical returns are lower than the previous policy. {\textbf{Mixup}} is a convex combination of the old and new parameters to smooth the transition and escape rejection deadlocks.
\textbf{Three-Points} evaluates an additional “midpoint” to choose the best of the old, new, and mixed policies. Requiring only \(\mathcal{O}(\epsilon^{-2}\ln(T/\delta))\) extra rollouts, it can be plugged into any PG method (DDPG, TRPO, PPO, etc.) without per-environment tuning. This design delivers theoretical improvement guarantees and consistent empirical gains in final performance, sample efficiency, and training stability. The framework can be related to line‐search and momentum  \citep{more1994line,sutskever2013importance,muehlebach2021optimization}, but is designed as a virtually zero-hyperparameter, plug‐in wrapper.

\section{Preliminaries}
\label{sec:prelim}
The foundational framework for Reinforcement Learning is the Markov Decision Process (MDP), roughly \(\Mcal=(\Scal,\Acal,P,r,\gamma,\rho)\), where \(\Scal\) is the state space,
\(\Acal\) the action set, \(P(s'\mid s,a)\) the transition kernel,
\(r(s,a)\!\in[0,R_{\max}]\) the bounded one‐step reward,
\(\gamma\in[0,1)\) the discount factor, and \(\rho\) the initial state
distribution. At each step, the agent observes the current state $s_t$ and takes an action $a_t$. The environment then transits the next state $s_{t+1}$ according to the transition kernel $P(\cdot|s_t,a_t)$, and receiving an immediate reward $r(s_t,a_t)$. A parameterized Markov policy is a mapping $\pi_\theta:\Scal\to\Delta(\Acal)$\footnote{$\Delta(\cdot)$ is the probability simplex over the space $\cdot$.} that captures the probability of taking actions under each state, for some parameter $\theta\in\Theta$. A policy $\pi_\theta$ can randomly induce a trajectory $\tau=(s_0,a_0,s_1,...)$ under the MDP, whose return is defined as the accumulated reward along the trajectory: $G(\tau)=\sum_{t=0}^\infty\gamma^t\,r(s_t,a_t).$ The value function of a policy $\pi_\theta$ is then defined as the expectation of returns: \(V^\pi(s)\triangleq \E[G(\tau)\mid s_0=s]\). The goal is to find a policy $\pi$, following which the agent can get the highest cumulative reward: 
\begin{align}
    \theta^*=\arg\max_{\theta\in\Theta} J(\theta), \text{ where } J(\theta)\triangleq \E[V^{\pi_\theta}(s)|s\sim \rho], 
\end{align}
for some initial distribution $\rho$. 
  
Policy gradient algorithms optimize $J(\theta)$ through gradient ascent, based on the Policy Gradient Theorem \citep{sutton1999policy}: $\nabla_\theta J(\theta)
  = \E_{\tau\sim\pi_\theta}\Bigl[\sum_{t=0}^\infty
      \nabla_\theta\log\pi_\theta(a_t\mid s_t)\,Q^{\pi_\theta}(s_t,a_t)
    \Bigr]$,
where the action‐value function is defined as 
\(Q^\pi(s,a)\triangleq \E[G(\tau)|s_0=s,a_0=a]\).  

In practice, as the value functions are unknown, one needs to replace them with the (Monte‐Carlo) estimation, resulting in the straightforward REINFORCE algorithm \citep{sutton1998reinforcement}. However, REINFORCE suffers from variances in gradient estimation and unstable performance; hence, modern approaches employ distinct stabilization mechanisms through constrained optimization and off-policy learning. Among on-policy methods, TRPO maximizes a surrogate‐advantage objective subject to a hard KL‐constraint \citep{schulman2015trust}:
\begin{equation*}
\begin{aligned}
\max_{\theta}\quad 
&\mathcal{L}^{\rm TRPO}(\theta)
=\;\mathbb{E}_{\tau\sim\pi_{\theta_{\rm old}}}\Bigl[
   r_t(\theta)\,\hat A^{\pi_{\theta_{\rm old}}}(s_t,a_t)
\Bigr],\\
\text{s.t.}\quad 
&\mathbb{E}_{\tau\sim\pi_{\theta_{\rm old}}}
\bigl[\mathbb{D}_{KL}\bigl(\pi_{\theta_{\rm old}}(\cdot\!\mid s_t)\,\Vert\,\pi_{\theta}(\cdot\!\mid s_t)\bigr)\bigr]
\le\delta,
\end{aligned}
\end{equation*}
where $r_t(\theta)=\frac{\pi_\theta(a_t\mid s_t)}{\pi_{\theta_{\rm old}}(a_t\mid s_t)}$, and $\hat A$ being an advantage estimator.
TRPO enjoys a theoretical guarantee of (approximate) monotonic policy improvement with the KL constraint. PPO replaces TRPO’s hard constraint with a clipped surrogate that is cheaper to compute, still discourages large updates \citep{schulman2017proximal}. Unlike TRPO’s KL‐based trust region, PPO does not admit the same theoretical assurance of monotonic improvement, yet a practical heuristic that works well in large‐scale implementations. Meanwhile, DDPG has remained a standard off-policy PG algorithm, training a deterministic actor \(\mu_\theta:\Scal\to\Acal\) and a critic \(Q_\phi\) off-policy using a replay buffer \(\mathcal{D}\) and slowly-updated target networks \citep{lillicrap2016continuous} \((\mu_{\theta'}, Q_{\phi'})\):
{\small
\begin{eqnarray*}
\mathcal{L}(\phi) 
&=&\E_{(s,a,r,s')\sim\Dcal}\Bigl[r + \gamma\,Q_{\phi'}\bigl(s',\mu_{\theta'}(s')\bigr) - Q_\phi(s,a)\Bigr]^2,\\
\nabla_\theta J 
&\approx&\E_{s\sim\Dcal}\Bigl[\nabla_a Q_\phi(s,a)\big\vert_{a=\mu_\theta(s)}\,\nabla_\theta \mu_\theta(s)\Bigr].
\end{eqnarray*}}
Twin Delayed DDPG (TD3) \citep{fujimoto18addressing} further stabilizes DDPG by (i) using two critics and taking the minimum to reduce overestimation bias, (ii) delaying policy and target network updates, and (iii) adding clipped noise to target actions for policy smoothing.

\section{Reward Profiling Framework}\label{sec:framework}
In its simplicity, our reward profiling framework compares the performance of two policies $\pi_{\theta_1}, \pi_{\theta_2}$. Within any PG method step, a potential update on the policy is made: $\theta_1\to\theta_2$. This updated, as mentioned, is based on the inaccurate gradient estimation, which is the key source of high variance, leading to occasional updates that \emph{decrease} the true performance \(J(\pi)\), slow overall convergence, and provide no safeguard on the quality of the policy selection. Under the reward profiling framework, this update contributes only if it improves the performance. In the ideal case, we exactly know the value functions of $J(\theta_1)$ and $J(\theta_2)$, then we will only accept the update $\theta_1\to \theta_2$ if $J(\theta_2)\geq J(\theta_1)$, which results in a monotonically increasing performance and stabilizes training. In practice, however, we do not know $J(\theta)$ and need to make the comparison based on estimation. It is clear that the performance and stability highly depend on the accuracy of the comparison. To handle this, we provide our \textbf{high-confidence comparison} scheme. 

In general, for two candidate parameterized policies  $\pi_j$ with $j \in \{1,2\}$, a number (\textit{E}) of \textit{i.i.d.} trajectories are sampled from each policy forming evaluation sample sets $\mathcal{D}_j = \{ \tau_i^{(j)} \}_{i=1}^E \sim \pi_j$. The empirical return estimate for $\pi_j$ is computed as
\begin{equation}\label{eq:J_hat}
\hat{J}(\pi_j) = \frac{1}{|\mathcal{D}_j|} \sum_{\tau \in \mathcal{D}_j} G(\tau),
\end{equation}where $G(\tau)$ denotes the cumulative (discounted) return of trajectory $\tau$. Based on the comparison of these return estimates, the corresponding parameters are selected for the policy. We refer to this scheme as the "\textit{lookback}" framework, where we compare the performance of the policies from two successive steps:
\begin{equation}
    \theta_{t+1}
  = \begin{cases}
    \theta_{\rm new}, & \hat{J}(\pi_{\rm new})\geq \hat{J}(\pi_{\rm old}),\\
    \theta_{\rm old}, & \text{otherwise}.
  \end{cases}
\end{equation}

If the noise masks genuine improvements in the noisy estimates, the agent might get stuck. To address this, we draw inspiration from the method \emph{mixup}, a classic data augmentation technique used in supervised learning \citep{zhang2017mixup}, where inputs and labels are convexly combined to reduce inductive bias, widely used in computer vision applications \citep{dutta2024multiview}. Specifically, we consider an intermediate policy defined by
\begin{equation}
  \theta_{\rm mix}
  = \lambda\,\theta_{\rm new} + (1-\lambda)\,\theta_{\rm old},  
\end{equation}
  
where $\lambda\in [0,1]$ is a mixing parameter. Similarly, we compare their performance and accept the update if it results from a higher reward: 
\begin{equation}
  \theta_{t+1}
  = \arg\max_{\theta\in\{\theta_t,\theta_{\rm mix}\}}
    \hat{J}(\pi_{\theta}).
\end{equation}
We also highlight that, as \(\theta_{\rm mix}\) lies closer to \(\theta_t\) in parameter space (when $\lambda$ is small), this step functions as a cheap trust region, yet without any Hessian computation or KL constraint as in TRPO \citep{schulman2015trust,schulman2017proximal}. This step is expected to result in a better performance.
Finally, the last modification unifies the best of the first two cases. The \emph{Three‐Points} variant considers all candidates \(\{\theta_t,\theta',\theta_{\rm mix}\}\) and selects the best:
\begin{equation}
  \theta_{t+1}
  = \arg\max_{\theta\in\{\theta_{\rm old},\theta_{\rm new},\theta_{\rm mix}\}}
    \hat{J}(\pi_{\theta}).
\end{equation}
By design, this technique ensures that $\pi$ mostly improves over the training process and provides enough room for exploration. We summarize the entire procedure in Algorithm \ref{alg:profiling-variants}, noting that it can be wrapped around any first‐order PG method without altering its core update logic.

\begin{algorithm}[tb]
\SetAlgoLined
\KwIn{initial \(\theta_0\); iterations \(T\); rollouts \(E\); mix‐weight \(\lambda\); variant \(\in\{\mathrm{LB,MU,TP}\}\)}
\For{\(t=0,\dots,T-1\)}{
  \texttt{Update the policy following any PG method to the new parameter} $\theta'$ \\
  \(\theta_{\rm mix}\!\leftarrow\!\lambda\,\theta' + (1-\lambda)\,\theta_t\)  \hfill (Eq.~(4))\\
  \texttt{Estimate }\(\hat{J}_{\rm old},\hat{J}_{\rm new},\hat{J}_{\rm mix}\)  \texttt{ with } \hfill \\
  \(\theta_{t+1}\!\leftarrow\)
  \(\displaystyle
     \begin{cases}
       \arg\max_{\{\theta_t,\theta'\}}\hat{J}, & \backslash\backslash{} \texttt{Lookback}\backslash\backslash{} \\
       \arg\max_{\{\theta_t,\theta_{\rm mix}\}}\hat{J}, & \backslash\backslash{} \texttt{Mix-up}\backslash\backslash{} \\
       \arg\max_{\{\theta_t,\theta',\theta_{\rm mix}\}}\hat{J}, & \backslash\backslash{}\texttt{Three-points}\backslash\backslash{}
     \end{cases}
  \)  \hfill (Eqs.~(5),(6))

}
\caption{Reward Profiling Framework \vspace{-2.5mm}
}
\label{alg:profiling-variants}
\end{algorithm}

\section{Convergence Analysis}\label{sec:convex case}
To provide theoretical validation of the framework, we study the choice of $E$, which serves as a sort of evaluation budget, in our method to ensure accuracy, and then we propose a detailed convergence analysis. We start with adapting the following concentration inequality.

\begin{table*}[ht]
\centering
\setlength{\tabcolsep}{1pt}
\small
\begin{tabular}{@{} ll *{4}{r} *{4}{r} *{3}{r} @{}}
\toprule
\multirow{2}{*}{Env.} & \multirow{2}{*}{Algo} & \multicolumn{4}{c}{Avg.\ Return ($\pm$ Std)} & \multicolumn{4}{c}{Rounds to 0.95×Best} & \multicolumn{3}{c}{Variability $\downarrow$ (\%)} \\
\cmidrule(lr){3-6} \cmidrule(lr){7-10} \cmidrule(lr){11-13}
 &  & Base & LB & MU & TP & Base & LB & MU & TP & LB & MU & TP \\
\midrule
Bipedal & DDPG & $-116.1 \pm 13.5$ & $-100.0 \pm 6.9$ & $\mathbf{-47.0 \pm 27.1}$ & $-58.9 \pm 34.4$ & -- & $\mathbf{1.0}$ & $1.3$ & $3.7$ & $-39$ & $\mathbf{-34}$ & $-64$ \\
 & PPO & $-51.9 \pm 109.2$ & $-3.1 \pm 66.1$ & $-13.4 \pm 8.0$ & $\mathbf{8.9 \pm 39.5}$ & $\mathbf{6.8}$ & $9.0$ & -- & $9.0$ & $66$ & $\mathbf{68}$ & $59$ \\
 & TRPO & $\mathbf{105.0 \pm 57.0}$ & $-35.3 \pm 11.1$ & $-17.9 \pm 7.2$ & $-27.9 \pm 16.1$ & $\mathbf{7.7}$ & -- & -- & -- & $67$ & $\mathbf{82}$ & $56$ \\
\midrule
CarRacing & DDPG & $-84.7 \pm 4.2$ & $\mathbf{-17.6 \pm 3.6}$ & $-17.6 \pm 3.6$ & $-17.6 \pm 3.6$ & -- & $\mathbf{1.6}$ & $3.6$ & $3.0$ & $-28$ & $\mathbf{-23}$ & $-26$ \\
 & PPO & $-63.2 \pm 14.3$ & $21.7 \pm 23.0$ & $26.9 \pm 21.8$ & $\mathbf{30.7 \pm 39.5}$ & $2.3$ & $3.0$ & $4.6$ & $\mathbf{2.0}$ & $\mathbf{-33}$ & $-48$ & $-45$ \\
 & TRPO & $-72.8 \pm 23.7$ & $-14.7 \pm 11.0$ & $31.1 \pm 45.2$ & $\mathbf{42.4 \pm 46.3}$ & $3.5$ & -- & $3.7$ & $\mathbf{3.0}$ & $\mathbf{57}$ & $-22$ & $-49$ \\
\midrule
LunarLander & DDPG & $-117.0 \pm 45.3$ & $-29.9 \pm 35.8$ & $26.3 \pm 98.4$ & $\mathbf{37.1 \pm 78.4}$ & -- & $\mathbf{1.0}$ & $6.7$ & $5.7$ & $12$ & $23$ & $\mathbf{27}$ \\
 & PPO & $\mathbf{-6.9 \pm 132.3}$ & $-19.0 \pm 113.4$ & $-71.9 \pm 87.3$ & $-40.6 \pm 42.5$ & $7.3$ & $\mathbf{6.7}$ & $9.0$ & $6.8$ & $5$ & $4$ & $\mathbf{28}$ \\
 & TRPO & $-15.0 \pm 89.0$ & $21.9 \pm 80.5$ & $79.0 \pm 93.2$ & $\mathbf{94.1 \pm 112.8}$ & $\mathbf{5.5}$ & $7.0$ & $8.7$ & $6.3$ & $41$ & $\mathbf{48}$ & $27$ \\
\midrule
Ant & DDPG & $140.6 \pm 113.9$ & $278.9 \pm 136.8$ & $\mathbf{412.4 \pm 150.0}$ & $377.7 \pm 115.4$ & $5.0$ & $\mathbf{4.0}$ & $4.2$ & $4.8$ & $3$ & $4$ & $\mathbf{11}$ \\
 & PPO & $822.6 \pm 35.8$ & $842.4 \pm 42.4$ & $768.3 \pm 97.3$ & $\mathbf{863.7 \pm 67.4}$ & $\mathbf{5.0}$ & $7.2$ & $9.0$ & $6.8$ & $\mathbf{39}$ & $-31$ & $19$ \\
 & TRPO & $\mathbf{897.6 \pm 41.2}$ & $808.8 \pm 62.0$ & $771.2 \pm 69.1$ & $837.6 \pm 47.1$ & $8.7$ & $8.0$ & $\mathbf{5.0}$ & $7.7$ & $20$ & $-67$ & $\mathbf{24}$ \\
\midrule
HalfCheetah & DDPG & $576.3 \pm 749.9$ & $909.0 \pm 331.4$ & $\mathbf{1137.8 \pm 185.6}$ & $931.0 \pm 345.4$ & $10.0$ & $\mathbf{7.0}$ & $8.6$ & $8.0$ & $43$ & $\mathbf{52}$ & $26$ \\
 & PPO & $\mathbf{97.1 \pm 377.9}$ & $34.2 \pm 596.5$ & $-1004.3 \pm 409.5$ & $95.5 \pm 561.8$ & $\mathbf{5.7}$ & $7.4$ & $10.0$ & $6.8$ & $-16$ & $\mathbf{48}$ & $-14$ \\
 & TRPO & $228.4 \pm 416.1$ & $81.2 \pm 524.3$ & $40.9 \pm 708.6$ & $\mathbf{391.9 \pm 314.9}$ & $7.3$ & $6.0$ & $7.0$ & $\mathbf{5.3}$ & $6$ & $-4$ & $\mathbf{17}$ \\
\midrule
Hopper & DDPG & $1189.6 \pm 600.4$ & $1048.7 \pm 555.5$ & $1394.5 \pm 727.7$ & $\mathbf{1492.2 \pm 562.9}$ & $7.8$ & $\mathbf{4.5}$ & $5.8$ & $6.3$ & $\mathbf{-3}$ & $-16$ & $-8$ \\
 & PPO & $\mathbf{825.5 \pm 71.4}$ & $802.8 \pm 49.6$ & $338.9 \pm 330.7$ & $634.3 \pm 281.6$ & $\mathbf{8.4}$ & $9.0$ & $9.0$ & $10.0$ & $-42$ & $\mathbf{-25}$ & $-88$ \\
 & TRPO & $\mathbf{1196.3 \pm 464.7}$ & $856.5 \pm 526.0$ & $646.5 \pm 283.6$ & $724.4 \pm 219.2$ & $\mathbf{8.2}$ & $9.0$ & -- & -- & $-20$ & $\mathbf{19}$ & $-8$ \\
\midrule
Walker2D & DDPG & $113.6 \pm 90.7$ & $\mathbf{299.4 \pm 74.8}$ & $248.7 \pm 145.6$ & $267.6 \pm 88.5$ & $8.0$ & $5.0$ & $\mathbf{4.9}$ & $5.4$ & $\mathbf{-41}$ & $-80$ & $-44$ \\
 & PPO & $185.6 \pm 107.2$ & $110.8 \pm 40.2$ & $\mathbf{199.3 \pm 108.2}$ & $126.4 \pm 25.1$ & $4.4$ & $1.8$ & $1.8$ & $\mathbf{1.2}$ & $-89$ & $-86$ & $\mathbf{-63}$ \\
 & TRPO & $\mathbf{260.2 \pm 207.4}$ & $192.6 \pm 133.2$ & $141.6 \pm 54.4$ & $231.4 \pm 182.4$ & $3.0$ & $\mathbf{2.2}$ & $2.4$ & $2.4$ & $-45$ & $\mathbf{-19}$ & $-56$ \\
\midrule
Humanoid & DDPG & $-120.1 \pm 26.7$ & $42.6 \pm 12.5$ & $51.6 \pm 14.3$ & $\mathbf{57.1 \pm 14.3}$ & -- & $3.7$ & $\mathbf{2.8}$ & $3.1$ & $\mathbf{50}$ & $45$ & $48$ \\
 & PPO & $\mathbf{58.0 \pm 7.1}$ & $48.2 \pm 6.9$ & $47.9 \pm 9.1$ & $48.3 \pm 7.5$ & $8.1$ & $1.9$ & $2.2$ & $\mathbf{1.1}$ & $3$ & $\mathbf{7}$ & $4$ \\
 & TRPO & $\mathbf{68.8 \pm 6.4}$ & $44.8 \pm 6.2$ & $47.8 \pm 6.9$ & $47.0 \pm 7.2$ & $8.0$ & $\mathbf{2.0}$ & $3.0$ & $3.0$ & $-21$ & $\mathbf{-14}$ & $-28$ \\
\bottomrule
\end{tabular}
\vspace{1mm}
\caption{Comparison across environments and algorithms. \emph{Avg.\ Return} reports final mean $\pm$ std over seeds; \emph{Rounds to 0.95×Best} counts iterations to reach 95\% of the best average return; \emph{Variability $\downarrow$ (\%)} measures relative reduction in return variance. \textbf{Bold} indicates best performance; dash (--) indicates not reached within budget. Abbreviations: Base (vanilla baseline), LB (Lookback), MU (MixUp), TP (Three-Points).}
\label{tab:combined}
\end{table*}

\begin{lemma}[Concentration]
\label{lem:concentration}
Let the policy $\pi$ have per-step rewards lie in $[0,R_{\max}]$, so that any trajectory return satisfies
\(
0 \;\le\; G(\tau) \;\le\; B
\quad\text{with}\quad
B \;=\;\frac{R_{\max}}{1-\gamma}.
\)
Then for any $\epsilon>0$,
\[
\mathbb{P}\!\bigl(\bigl|\hat{J}(\pi)-J(\pi)\bigr|\ge\epsilon\bigr)
\;\le\;
2 \exp\!\Bigl(-\tfrac{2\,E\,\epsilon^2}{B^2}\Bigr).
\]
\end{lemma}

Applying Lemma~\ref{lem:concentration} with a per-step failure probability of $\tfrac{\delta}{T}$ and then union‐bounding over $T$ consecutive updates, we get for every $t \in [T]$ empirical estimates satisfies
\(\bigl|\hat{J}(\pi_t)-J(\pi_t)\bigr|\le\epsilon\),
provided $E \;\ge\;
\frac{B^2}{2\,\epsilon^2}\,
\ln\!\Bigl(\tfrac{2\,T}{\delta}\Bigr)$, 
with probability at least $1-\delta$. \footnote{This requirement is a worst-case bound on the return range. In our experiments, the variance observed is much lower with $E=5-10,$ only.}

\begin{lemma}[Monotonicity]\label{lem:mono-hp}

Let  
  $E \;\ge\; \frac{B^2}{2\,\epsilon^2}\,\ln\!\Bigl(\frac{2\,T}{\delta}\Bigr)$.
Then with probability at least \(1-\delta\), for every update \(t=1,2,\dots,T\), whenever the lookback rule accepts (i.e.\ \(\hat{J}(\pi_{\theta_{t+1}})\ge\hat{J}(\pi_{\theta_t})\)), the true returns satisfy
\[
  J\bigl(\pi_{\theta_{t+1}}\bigr)
  \;\ge\;
  J\bigl(\pi_{\theta_t}\bigr)
  \;-\;2\,\epsilon.
\]
\end{lemma}
This implies that, with this choice of $E$, the comparison ensures the accepted updates yield performance improvements, and hence the learning can be monotonic and more stable. 

We then show that when instantiated with the \emph{Lookback} decision rule and REINFORCE algorithm, Algorithm \ref{alg:profiling-variants} converges to a near-optimal policy at rate \(\mathcal{O}(T^{-1/4})\) with high probability. The analysis for the mix-up and three-point strategies follows similar lines. We make the following standard assumption.
\begin{assumption}[Smoothness]\label{ass:smooth}
The policy class \(\pi_\theta\) is smooth in \(\theta\), and there exists \(\sigma\ge0\) such that
\[
  \bigl\|\E_{d^{\pi_\theta}\times\pi_\theta}\bigl[(\nabla\log\pi_\theta(A|S))(\nabla\log\pi_\theta(A|S))^\top\bigr]\bigr\|\le\sigma,
\] 
where $d^{\pi_\theta}(s)$ is the discounted-state-visitation distribution under $\pi_{\theta}$. 
\end{assumption}
Assumption \ref{ass:smooth} is standard in policy gradient studies, e.g., \citep{xu2020improved, wang2024non, ganesh2024accelerated}. There is a rich family of policies that satisfy the conditions, including the Softmax policy and policies defined through neural networks with smooth and Lipschitz activation functions
\citep{tian2023convergence}.
Finally, we are all set to quote the convergence of our algorithm. 
\begin{theorem}[Convergence]\label{thm:conv-main}
Under Assumption \ref{ass:smooth}, setting \(\eta=\mathcal{O}(T^{-1/2})\) and $  E = \frac{B^2}{2\epsilon^2}\,\ln\!\Bigl(\tfrac{2T}{\delta}\Bigr),$ results in $J(\pi^*) - J(\pi_{\theta_T})= \mathcal{O}\bigl(T^{-1/4}\bigr),$ with probability at least \(1-\delta\).
\end{theorem}
 
The deterministic PG method \citep{agarwal2021theory, xiao2022convergence} achieves a faster convergence rate than ours because it has access to the true policy gradient without error. In contrast, our algorithm achieves a similar convergence rate when the algorithm is stochastic and the update is based on policy gradient estimation, as in actor-critic algorithms \citep{xu2020improved, suttle2023beyond, olshevsky2023smallgainanalysissingle, chen2021closing, wang2024non}. This demonstrates that the convergence rate of our approach is not slowed down much by our profiling framework. More importantly, unlike previous works, Theorem \ref{thm:conv-main} provides a high‐probability guarantee on the \emph{last} iteration, whereas prior results typically only guarantee the expectation or the best policy found during the learning process. This framework requires additional trajectories of order \(\mathcal{O}(\epsilon^{-2}\ln(T/\delta))\), which is still acceptable when the training samples are redundant.

\begin{remark}
    The same convergence guarantee of Theorem \ref{thm:conv-main} holds for the other two cases. We note that the Mixup is a special case of Lookback. Because $\theta_{\rm mix}=\lambda \theta'+(1-\lambda)\theta_t=\theta_t+\lambda\eta \hat\nabla_\theta J(\pi_{\theta_t}).$ Three points strategy adds another point in Lookback and generates a better reward than Lookback in every iteration. 
\end{remark}

\smartparagraph{Biased Case.} In the above discussions, we consider the case where we compare performances based on Monte-Carlo estimation. In practice, however, it is often the case that PG methods are utilized in \emph{actor-critic} algorithms, where the value functions are estimated based on the \emph{critic}. Since the \emph{critic} part is generally inaccurate, the resulting gradient estimations can have bias. Thus, it is important to extend our framework to the biased case. Given $\theta$, we assume that the critic part provides a biased estimation of the value function $\hat{Q}^{\pi_\theta}$, such that $\mathbb{E}[\hat{Q}^{\pi_{\theta}}] = {Q}^{\pi_{\theta}} + \epsilon_{\theta},$ where $\epsilon_{\theta}$ is the bias introduced by the errors of the critic part. Such results can be obtained, for instance, when a deep neural network is used to estimate value functions, e.g., \cite{du2019gradient,neyshabur2017implicit,miyato2018spectral}, and is widely adapted in actor-critic analysis, e.g., \citep{wang2019neural,zhou2023single,chen2023global,zhang2020provably,qiu2021finite,kumar2023sample,xu2020non,xu2020improving,suttle2023beyond}. It can be further shown that when the value function estimation is biased, the resulting gradient estimation has a bias of $CA\epsilon_\theta$, as long as $\|\nabla \pi_\theta\|\leq C$  \citep{sutton1999policy} and $A=|\mathcal{A}|$ actions. Therefore, without loss of generality, we assume the bias of zero-th and first order gradient of $Q^{\pi_\theta}$ are both bounded by some $\epsilon_\theta$. We then show the convergence of \ref{alg:profiling-variants}, when the Lookback technique is applied to REINFORCE.

\begin{theorem}\label{thm:convergence}
We set $\eta = \mathcal{O}(1/\sqrt{T})$ and run \ref{alg:profiling-variants}(Lookback) for $T$ steps. If the bias $\epsilon_\theta$ satisfies $\epsilon_\theta \leq \mathcal{O}(1/T)$, then we have  $\min_{t\leq T}\min\{\mathbb{E} \|\nabla {V}^{\pi_{\theta_t}}\|_{2}, \mathbb{E} \|\nabla {V}^{\pi_{\theta_t}}\|_{2}^2 \} \leq \mathcal{O}\left(\frac{1}{\sqrt{T}}\right).$
\end{theorem}
The result shows that our algorithm converges to a stationary point, as long as the value estimation is accurate enough. Moreover, due to the gradient dominance property of the value function \citep{agarwal2021theory}, our result also implies the global optimality of the learned policy. 

\begin{figure*}[ht]
  \centering
  \includegraphics[width=\linewidth]
{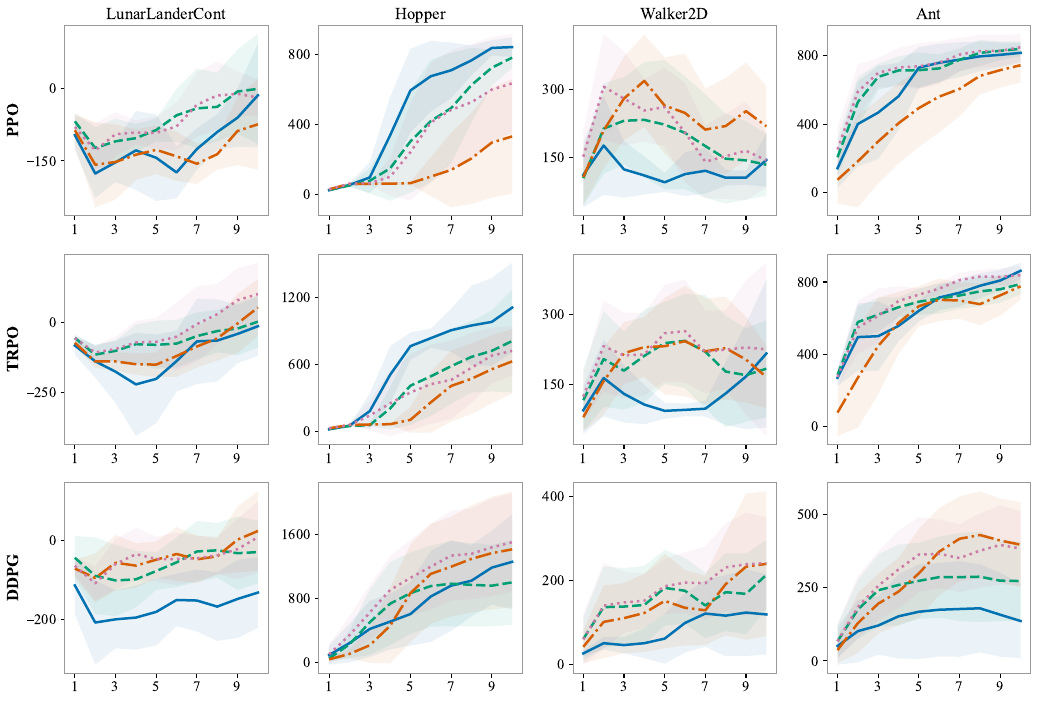}
  
  \caption{Average Return vs. timesteps ( $\times10^4$) for 3 policy-gradient 
  methods, PPO (top row), TRPO (middle row), and DDPG (bottom row), across 
  4 continuous-control benchmarks: LunarLanderCont, Hopper, Walker2D, Ant under variants: 
  {\color{MPLBlue}\textbf{——}} Vanilla, 
  {\color{MPLGreen}\textbf{– – –}} Lookback, 
  {\color{MPLOrange}\textbf{–·–·–}} Mixup, 
  {\color{MPLPurple}\textbf{·····}} Three-Points, 
  with $E = 5$ for minimal overhead. Shaded regions indicate $\pm 1$ standard 
  deviation. Extra rollouts tuning yields convergence gains in complex 
  environments.}
  \label{fig:all_return_curves}
\end{figure*}

\section{Experiments}\label{sec:experiments}
\subsection{Simulated Gymnasium Environments}
We benchmarked our reward-profiling wrapper by performing extensive experiments to stress-test policy learning across the major flavors of policy-gradient methods. We wrapped three canonical algorithms.
\textbf{TRPO}, the trusted “second-order” method with hard KL-constraints,
\textbf{PPO}, its “first-order” surrogate-clipping successor and current on-policy workhorse, and
\textbf{DDPG}, an off-policy, replay buffer actor-critic known for sample-efficient continuous control but notoriously unstable, in our profiling layer.

\smartparagraph{Profiling framework integration.} 
We built a modular profiling wrapper atop Stable-Baselines3 (v1.9) and SB3-Contrib \citep{stable-baselines3}, leveraging Gymnasium and PyBullet~\cite{benelot2018, towers2024gymnasiumstandardinterfacereinforcement} for simulation. Each variant ran for 100k environment steps across five random seeds. Each profiling round consumes the same $10$k environment steps as the baseline; samples used for improvement checks are \emph{reused} for training updates. Thus, no additional interactions are collected. All variants share the same per-round evaluation schedule and count. Every candidate set is scored with short episodes ($E=5$), and averages are computed on the identical rollouts for all baselines for fair comparison.
The policy network in SB3 is actor-critic, which was used throughout the experiments.

\smartparagraph{Benchmark environments.}
We evaluate reward profiling on two complementary families of continuous control tasks:
\textbf{Box2D Suite} where the BipedalWalker environment challenges agents to navigate uneven terrain using 24D lidar/joint observations and 4D torque actions, with sparse rewards for forward progress. CarRacing provides pixel-based control (96×96 RGB input) for lap completion, while LunarLanderContinuous tests precise thruster control (8D state, 2D actions) under sparse landing rewards. \textbf{MuJoCo/PyBullet Suite}, where we experimented with high-dimensional locomotion with Ant, HalfCheetah, Hopper, Humanoid, Walker2D, with observation state and action both having several dimensions, ranging from torque control, actions for balance, or even for forward gait.   

\begin{figure}[t]
    \centering
    \setlength{\fboxsep}{1pt}
    \fbox{\includegraphics[width=\columnwidth]{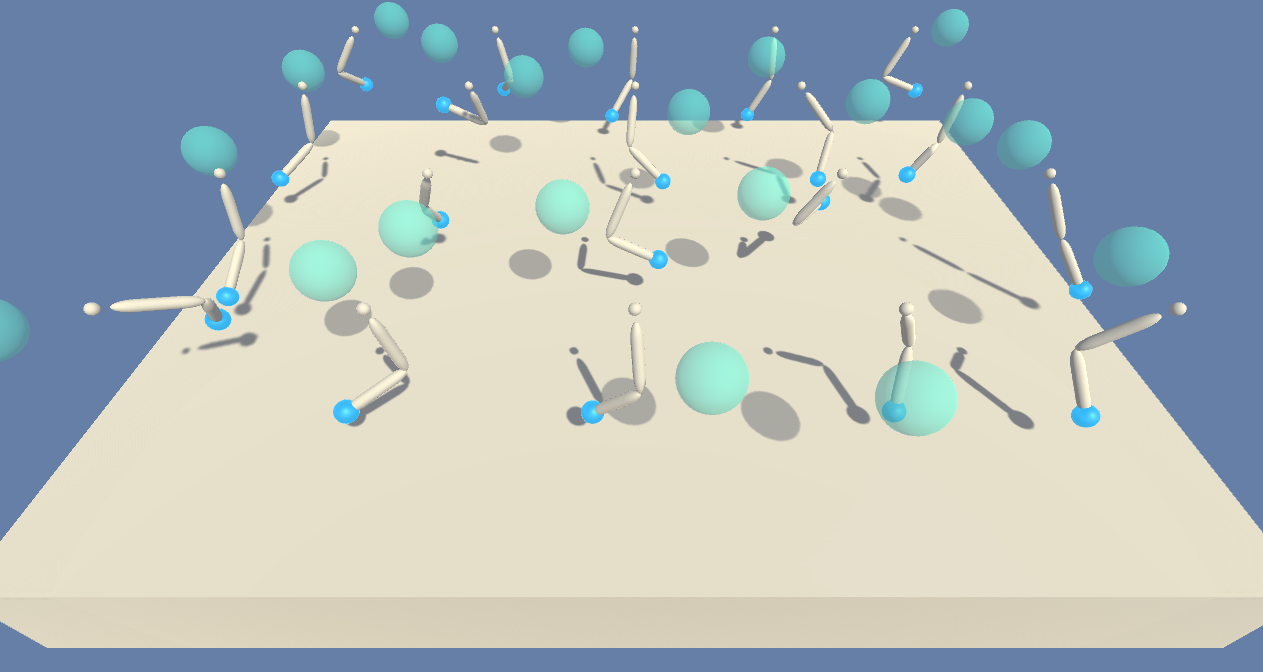}}
    \caption{Multi-agent Reacher task in UnityML: multiple arms coordinate to reach and manipulate randomly placed targets on a raised platform, testing control under interaction.}
    \label{fig:reacher-scene}
\end{figure}
\begin{figure}[t]
    \centering
    \includegraphics[width=\columnwidth]{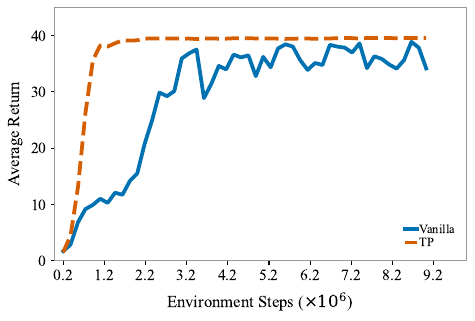}
    \caption{Perormance in the UnityML Reacher environment. Profiling-enhanced DDPG converges faster and maintains greater stability than the vanilla counterpart.}
    \label{fig:reacher-curve}
\end{figure}

\smartparagraph{Observation.} The most significant gain is in terms of variance performance overall as shown in Figure \ref{fig:all_return_curves}. In cases it transforms catastrophic failures into viable policies across key benchmarks. The CarRacing agent, both with TRPO and PPO, average large negative returns, crashing before completing a lap. The variants turn them into positive‐return drivers. The training recovers stable driving behavior by modifying updates that worsen lap scores. DDPG remains unstable under an MLP policy, but profiling still raises the worst runs. For BipedalWalker, where vanilla TRPO already performs well ($105.0 \pm 57.0$), profiling yields minimal gains due to inherent conservatism, though DDPG$+$Mixup still improves from $-116.1 \pm 13.5$ to $-47.0 \pm 27.1$. It still provides competitive performance in precision tasks. Among our high-DOF benchmarks, Ant DDPG$+$Mixup gives modest bumps over the baseline with earlier convergence behavior, while its Three-Point variant provides superior variance characteristics (11\% reduction from the baseline) with moderate final return over time. It also augments the training behavior with TRPO and PPO on several metrics, offering moderate improvement in variance. On HalfCheetah, DDPG$+$Mixup dominates ($1137.8 \pm 185.6$ vs $576.3 \pm 749.9$) with usual variance performance in PPO. The performance on Walker2D and Humanoid remain challenging in variance stability; even with profiling, TRPO and PPO show limited gains while baselines also show very similar results. The profiling framework improves most metrics with strong to moderate gains, except for the setups that remain challenging to learn with off-the-shelf algorithms without careful tuning. 

\subsection{Experiment with Multi-robot Task}

For a realistic and domain-specific evaluation, we deployed it in a multi-agent robotic setting on the Unity ML-Agents Reacher task \citep{juliani2020, cohen2022}. Unity provides a flexible, visually realistic, and customizable simulation environment; see Figure \ref{fig:reacher-scene}. Here, the environment simulates 20 robotic arms operating in parallel, each receiving a 33-dimensional state vector (joint angles, velocities, and target position) and outputting 4D continuous torques. Agents earn a dense reward of $+0.1$ per timestep for maintaining their fingertip in a moving target zone. We plugged our profiling wrapper (the TP variant) into the platform atop a decentralized DDPG architecture, with policy updates evaluated every 50 episodes through 5-agent consensus voting.


\begin{figure}[t]
    \centering
    \includegraphics[width=\columnwidth]
    {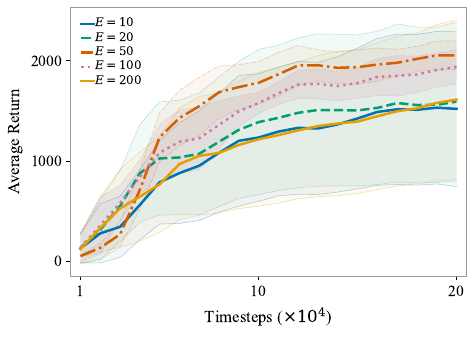}
    \caption{Sensitivity to evals \(E\) for DDPG+TP on Hopper. Smaller \(E\)(10–20) causes noisy, unstable updates; large \(E\)(200) improves stability but slows progress; mid-range \(E\)(50–100) balances both.}
    \label{fig:eval-budget}
\end{figure}

\smartparagraph{Observation.}~Three-Point profiling (TP) demonstrates substantially more stable learning than vanilla DDPG, as evidenced by the training performance in Figure~\ref{fig:reacher-curve}. TP-enhanced agent achieves a higher return from early training with fewer fluctuations, achieving smoother progression. Only five sampled returns (\(E=5\)) make the computation overhead minimal while solving for the optimal policy at early training iterations. This implies the suitability of a reward-profiling framework for physical deployment, where stable, predictable learning behavior is a requirement.

\section{Evaluation Rollout Sensitivity}

Using an NVIDIA V100 GPU, baseline training of 100K steps took 10–30 min, with the TP variant adding only about 20\% wall-clock time. Figure~\ref{fig:eval-budget} shows how the evaluation budget shapes learning.  At small \(E\) (10-20), the Monte Carlo estimates exhibit noisy, spurious updates and wide confidence bands. This results in large fluctuations and wide confidence bands.  At the other extreme, very large \(E\) (e.g., 200), the algorithm advances only slowly once the noise floor is suppressed.  Intermediate values (\(E=50\)–100) suppress the worst of the noise without over‑constraining the update rule, yielding both rapid early gains and a much narrower variance envelope. So roughly, pushing \(E\) well beyond this critical scale offers diminishing returns. Once the estimation error is negligible, further rollouts merely add computational cost and slow down true policy updates. This reflects the batch-size trade‑off in SGD \cite{mccandlish2018empirical}, where a task‑dependent \emph{critical batch size} marks the point beyond which additional samples no longer improve data efficiency.  By analogy, one can think of a \emph{critical evaluation budget} \(E_{\rm crit}\), and choose \(E\approx E_{\rm crit}\), to optimally balance variance reduction against responsiveness. Our framework complements other recent monotonic improvement goals~\cite{xie2025spo} without relying on stochastic density ratios, hence applies uniformly to both stochastic and deterministic settings.

\section{Conclusion and Open Challenges}

We introduced \textit{Reward Profiling}, a general-purpose wrapper applicable to policy gradient-based methods to enforce monotonic improvements. By ``looking back'' at the pre-update performance after gradient backpropagation and conditionally accepting, rejecting, or blending candidate updates, catastrophic collapses are reduced through stable learning. Empirically, across both on-policy (\textbf{PPO}, \textbf{TRPO}) and off-policy (\textbf{DDPG}) algorithms with actor-critic frameworks, and across dense-reward control tasks and high-DOF locomotion benchmarks, Reward Profiling mostly matched or improved returns along with the variance, sped up convergence relative to vanilla baselines. A key trade-off governed by the number of evaluation rollouts $E$ for practical implementation: moderate values of $E$ tend to strike the best balance between stability and responsiveness. Watching DDPG's triumph with it, the integration of reward profiling into the \texttt{Reacher} task demonstrates its effectiveness in a realistic continuous-control setting. Profiling yielded markedly smoother learning curves under DDPG, motivating its use in off-policy, replay-buffer cases. While Reward Profiling incurs only $\mathcal{O}(\log T)$  rollouts over $T$ iterations, the cost may be nontrivial in expensive simulators. On sparse-reward and large discrete-action tasks (e.g., Atari), this could be tested, adjusting $E$ dynamically based on some uncertainty measure, or incorporating more selective updates for the most informative candidates.

\section*{Acknowledgments}
The work of S.A and Y.W is partially supported by DARPA under Agreement No. HR0011-24-9-0427.

\bibliography{aaai2026cr}



\clearpage
\clearpage
\onecolumn
\appendix
\setcounter{secnumdepth}{1} 
\renewcommand\thesection{\Alph{section}}
\section*{APPENDIX}

\smartparagraph{Organization.}
We organize the Appendix into two main parts. \textbf{Section~\ref{appendix:theory}} presents the theoretical materials used in the main text and provide proofs of concentration bounds, monotonicity guarantees, smoothness results, and the
convergence theorems stated in \S\ref{sec:convex case} of the main paper. \textbf{Section~\ref{appendix:impl}} contains implementation and experimental details.  
We mention the network architectures, hyperparameters, and profiling variants,
followed by more empirical results, including PPO/TRPO/DDPG/TD3 ablations,
full return trajectories across all environments, and the Unity ML-Agents
Reacher setup and integration details reported in \S\ref{sec:experiments}.



\section{Theoretical Guarantee}\label{appendix:theory}

This appendix contains complete proofs of all theoretical claims in  \S\ref{sec:convex case}.

\subsection{A.1 \; Concentration Inequality (Proof of Lemma~\ref{lem:concentration})}
\label{appendix:concentration}
Let $\{\tau_i\}_{i=1}^E$ be the $E$ \textit{i.i.d.}\ trajectories sampled from policy $\pi$, and define
\[
X_i \;\coloneqq\; G(\tau_i),
\]
so that by construction
\[
0 \;\le\; X_i \;\le\; B,
\qquad
\E[X_i] \;=\; J(\pi),
\]
where $B = R_{\max}/(1-\gamma)$ and $J(\pi)=\E_{\tau\sim\pi}[G(\tau)]$ is the true expected return.  From Eq.~\eqref{eq:J_hat} we have
\[
\hat J(\pi) = \mathbb{E}_E[X]=\;\tfrac{1}{E}\sum_{i=1}^E X_i.
\]

Applying Hoeffding’s inequality to the bounded, independent variables $\{X_i\}$ \citep{hoeffding1994probability} gives for any $\epsilon>0$ the one‑sided tail bounds
\[
\mathbb{P}\!\Bigl(\hat{J}(\pi) - J(\pi)\ge\epsilon\Bigr)
\;\le\;
\exp\!\Bigl(-\tfrac{2\,E\,\epsilon^2}{B^2}\Bigr),
\]
A union bound yields
\[
\mathbb{P}\!\Bigl(\bigl|\hat{J}(\pi)-J(\pi)\bigr|\ge\epsilon\Bigr)
\;\le\;
2\exp\!\Bigl(-\tfrac{2\,E\,\epsilon^2}{B^2}\Bigr),
\]
as claimed. \qed


\subsection{A.2 \; High‐Probability Monotonicity (Proof of Lemma~\ref{lem:mono-hp})}
\label{appendix:mono-hp}

From Lemma~\ref{lem:concentration}, each empirical estimate \(\hat J(\pi)\) satisfies
\[
  \mathbb{P}\bigl(|\hat J(\pi)-J(\pi)|\le\epsilon\bigr)
  \;\ge\;
  1 - 2\exp\!\Bigl(-\tfrac{2E\,\epsilon^2}{B^2}\Bigr).
\]
By our choice of \(2\exp(-2E\epsilon^2/B^2)\le\delta/T\), each single evaluation errs by at most \(\epsilon\) with prob. \(\ge1-\delta/T\).  At each update \(t\) two independent evaluations (the “old” policy, and the “new”) are performed, so by a union bound over the \(T\) updates, with prob. \(\ge1-\delta\) \emph{all} \(2T\) estimates are simultaneously within \(\epsilon\) of their true values.
Let's fix any iteration $t\in[T]$ at which the lookback rule accepts:
\[
  \hat J\bigl(\pi_{\theta_{t+1}}\bigr)
  \;\ge\;
  \hat J\bigl(\pi_{\theta_t}\bigr).
\]
Since both estimates deviate from the truth by at most \(\epsilon\), we have
\[
  J\bigl(\pi_{\theta_{t+1}}\bigr)
    \;\ge\;
  \hat J\bigl(\pi_{\theta_{t+1}}\bigr) - \epsilon
    \;\ge\;
  \hat J\bigl(\pi_{\theta_t}\bigr) - \epsilon
    \;\ge\;
  J\bigl(\pi_{\theta_t}\bigr) - 2\,\epsilon.
\]
Since this holds for every \(t\in\{1,\dots,T\}\) with prob. \(1-\delta\), the lemma follows. \qed

\subsection{A.3 \; Verification of Assumption~\ref{ass:smooth} (Softmax Policy Class)}
\label{appendix:softmax}
\begin{lemma}

If  
\[
\pi_\theta(a\mid s)\propto\exp\bigl(-\phi(s,a)^\top\theta\bigr)
\quad\text{with}\quad
\|\phi(s,a)\|\le1,
\]
then for all $(s,a)$,
\[
\bigl\|\nabla_\theta\log\pi_\theta(a\mid s)\bigr\|\le2.
\]

\end{lemma}
We have
\[
  \log\pi_\theta(a\mid s)
  = -\,\phi(s,a)^\top\theta
    - \log\!\sum_{b}e^{-\phi(s,b)^\top\theta}.
\]
Hence
\[
  \nabla_\theta\log\pi_\theta(a\mid s)
  = -\,\phi(s,a)
    + \frac{\sum_{b}e^{-\phi(s,b)^\top\theta}\,\phi(s,b)}
           {\sum_{b}e^{-\phi(s,b)^\top\theta}}.
\]
By the triangle inequality and \(\|\phi(s,\cdot)\|\le1\),
\(\|\nabla\log\pi_\theta(a\mid s)\|\le1+1=2.\) \qed

\subsection{A.4 \; Smoothness and Descent Properties}\label{appendix:descent}


\begin{lemma}[Smoothness of $J(\theta)$]\label{lem:smooth-V}
Under Assumption~\ref{ass:smooth}, the expected return $J(\pi_\theta)$ is $L$‑smooth in $\theta$: for all $\theta,\theta'$,
\[
\|\nabla J(\pi_\theta)-\nabla J(\pi_{\theta'})\|\le L\,\|\theta-\theta'\|.
\]
\end{lemma}

Follows from standard policy gradient smoothness results (e.g.\ Lemma D.3 in \citep{agarwal2021theory}), using the bounded Fisher information in Assumption \ref{ass:smooth}. \qed


\begin{lemma}[Descent step]\label{lem:descent}
Let $G_t$ be an unbiased estimate of $\nabla J(\pi_{\theta_t})$ with $\|G_t\|\le G_{\max}$.  If
\(
\theta_{t+1} = \theta_t + \eta\,G_t,
\)
then
\[
\E\bigl[J(\pi_{\theta_{t+1}})\mid\theta_t\bigr]
\;\ge\;
J(\pi_{\theta_t})
\;+\;\eta\,\|\nabla J(\pi_{\theta_t})\|^2
\;-\;\tfrac{L}{2}\,\eta^2\,G_{\max}^2.
\]
\end{lemma}


By the $L$‑smoothness of $J(\pi_\theta)$ (Lemma~\ref{lem:smooth-V}), for any $\theta,\theta'$ we have
\[
J\bigl(\pi_{\theta'}\bigr)
\;\ge\;
J\bigl(\pi_{\theta}\bigr)
\;+\;\nabla J(\pi_{\theta})^\top(\theta'-\theta)
\;-\;\frac{L}{2}\,\|\theta'-\theta\|^2.
\]
Setting $\theta = \theta_t$ and $\theta' = \theta_{t+1} = \theta_t + \eta\,G_t$ gives
\[
J\bigl(\pi_{\theta_{t+1}}\bigr)
\;\ge\;
J\bigl(\pi_{\theta_t}\bigr)
\;+\;\eta\,\nabla J(\pi_{\theta_t})^\top G_t
\;-\;\frac{L}{2}\,\eta^2\,\|G_t\|^2.
\]
Taking the conditional expectation $\E[\cdot\mid\theta_t]$ and using linearity of expectation, the unbiasedness $\E[G_t\mid\theta_t]=\nabla J(\pi_{\theta_t})$, and the bound $\|G_t\|\le G_{\max}$, we obtain
\[
\begin{aligned}
\E\bigl[J(\pi_{\theta_{t+1}})\mid\theta_t\bigr]
&\ge
J\bigl(\pi_{\theta_t}\bigr)
\;+\;\eta\,\nabla J(\pi_{\theta_t})^\top\E[G_t\mid\theta_t]
\;-\;\frac{L}{2}\,\eta^2\,\E\bigl[\|G_t\|^2\mid\theta_t\bigr]\\
&=
J\bigl(\pi_{\theta_t}\bigr)
\;+\;\eta\,\|\nabla J(\pi_{\theta_t})\|^2
\;-\;\frac{L}{2}\,\eta^2\,G_{\max}^2,
\end{aligned}
\]
which yields the claimed descent inequality. \qed

\subsection{A.5 \; Proof of Theorem \ref{thm:conv-main}}
\label{appendix:proof-main}
Choose 
\(\eta=O(T^{-1/2})\), \(\epsilon=O(T^{-1/4})\), and 
\(E=\frac{B^2}{2\epsilon^2}\ln\!\bigl(\tfrac{2T}{\delta}\bigr)\). 

By Lemma~\ref{lem:descent} and telescoping,
\[
\frac1T\sum_{t=0}^{T-1}\E\|\nabla J(\pi_{\theta_t})\|^2
=O(T^{-1/2}),
\]
so \(\min_{t<T}\|\nabla J(\pi_{\theta_t})\|=O(T^{-1/4})\).  Lemma 4\.1 of\cite{agarwal2021theory} then gives
\[
J(\pi^*) - \max_{t<T}J(\pi_{\theta_t})
=O(T^{-1/4}).
\]

Meanwhile, Lemma~\ref{lem:mono-hp} ensures (w.p.\(\ge1-\delta\)) each accepted update loses at most \(2\epsilon=O(T^{-1/4})\), so, by immediately reverting any worse update, the final policy's return can never fall more than \(2\epsilon\) below the best one seen:
\[
\max_{t<T}J(\pi_{\theta_t})\;\le\;J(\pi_{\theta_T})+2\epsilon.
\]

Combining the two,
\[
J(\pi^*) - J(\pi_{\theta_T})
=O(T^{-1/4}),
\]
as claimed. \qed

\subsection{A.6 \; Proof of Theorem~\ref{thm:convergence} (Biased Critic Case)}
\label{appendix:biased-proof}

In this section, we assume 
$$ \mathbb{E}[\tilde{V}_{t}] = {V}_{t} + \epsilon_t,$$
where $\epsilon_t$ is the bias that we assume is also upper bounded as well as its gradient by some $\epsilon \ge 0$, i.e., the estimation of the value function is accurate in both zero$^{\rm th}$ and first order. Such an assumption can be satisfied when a deep neural network is used to estimate value functions, e.g.,  \cite{du2019gradient,neyshabur2017implicit,miyato2018spectral}, and is widely assumed in actor-critic analysis, e.g.,  \citep{wang2019neural,zhou2023single,chen2023global,zhang2020provably,qiu2021finite,kumar2023sample,xu2020non,xu2020improving,suttle2023beyond}.

\begin{lemma}\label{lem:monotone1} Algorithm \ref{alg:profiling-variants} produces an  monotonic sequence of iterates modulo $\epsilon$. In the sense that ${V}_{t+1}\geq {V}_t - 2 \epsilon$ given $\theta_t$ and $G_t$ for all $k\geq 0$. 

\end{lemma}

We have by \ref{alg:profiling-variants}:
$
 \tilde{V}_{t+1}  \ge \tilde{V}_{t}. 
$ 
Now, by taking the conditional expectation of the previous inequality, conditional on $\theta_t$ and $G_t$, we get $
 {V}_{t+1} +\epsilon_{t+1} \ge {V}_{t} + \epsilon_t. 
$ which implies 
$
 {V}_{t+1}  \ge {V}_{t} - 2 \epsilon. 
$
 

%

\begin{lemma}\label{lemma: l-smooth1}(Lemma D.3 of \cite{agarwal2021theory})
The value functions are $L$-smooth,  i.e., there exists some constant $L$, such that
\begin{align}
    \|\nabla_\theta \tilde{V}_1-\nabla_\theta \tilde{V}_2\|\leq  L\|\theta_1-\theta_2\|. 
\end{align}
\end{lemma}

\begin{lemma}\label{lem:descent1} 
For all $k\geq 0$,
\begin{align}\label{eq:s88ss1}
\mathbb{E} \left[ V_{t+1}| \theta_t \right] \geq V_t +  \eta_t \|\nabla V_t\|_{2}^2 - \frac{L}{2}\eta_t^2 \mathbb{E} (\|G_t\|_{2}^2|\theta_t) - 2 \epsilon -\eta_t \epsilon \|\nabla V_t\|_{2}.
\end{align}
\end{lemma}

We have \begin{align*}
 \tilde{V}_{t+1}  \ge \tilde{V}(\theta_t + \eta G_t) \ge  \tilde{V}_{t} +  \eta_t G_t^T \nabla \tilde{V}_t - \frac{L}{2}\eta_t^2 \|G_t\|_{2}^2,  
\end{align*}
where the algorithm enforces the first inequality, and the second inequality comes from the $L$-Smoothness of $\tilde{V}$.

Now, by taking the conditional expectation of the previous inequality, conditional on $\theta_t$ and $G_t$, we get 
\begin{align*}
 {V}_{t+1} + {\epsilon}_{t+1}  \ge  {V}_{t} + {\epsilon}_{t}+  \eta_t G_t^T \nabla {V}_t + \eta_t G_t^T \nabla {\epsilon}_t - \frac{L}{2}\eta_t^2 \|G_t\|_{2}^2.
\end{align*}

Now, by taking the expectation over the randomness on $G_t$ and the upper bound on $\epsilon_t$ we get:
\begin{align*}
\mathbb{E} \left[ V_{t+1}| \theta_t \right] \geq V_t +  \eta_t \|\nabla V_t\|_{2}^2 - \frac{L}{2}\eta_t^2 \mathbb{E} (\|G_t\|_{2}^2|\theta_t) - 2 \epsilon -\eta_t \epsilon \|\nabla V_t\|_{2}.
\end{align*}


Before proving our main result, we need another intermediate result as stated below. 

\begin{lemma}\label{lem:bounded variance}
It holds that 
\begin{eqnarray}\label{eq:boundedVar}
\textstyle \mathbb{E}\left(\|G_t\|^2 | \theta_t \right) \leq \frac{\sigma}{(1-\gamma)^4}. 
\end{eqnarray}
\end{lemma}
Note that when $\theta_t$ is fixed, the stochastic policy gradient is \begin{align}
    G_t=\frac{1}{1-\gamma} \nabla \log \pi_t(a|s)Q_t(s,a),
\end{align}
where $(s,a)\sim d^\pi_t(S)\times \pi_t(A|S)$. Note that \begin{align}
    \|G_t\|^2&=\|\frac{1}{1-\gamma} \nabla \log \pi_t(a|s)Q_t(s,a) \|^2\nonumber\\
    &= \frac{1}{(1-\gamma)^2} Q_t(s,a)^2 \|\nabla \log\pi_t(a|s)\|^2\nonumber\\
    &\leq \frac{1}{(1-\gamma)^4} \|\nabla \log\pi_t(a|s)\|^2.
\end{align}
Now taking expectation w.r.t. $(s,a)$ implies that
\begin{align}
    \mE[\|G_t\|^2 | \theta_t]&=\frac{1}{(1-\gamma)^4}  \mE_{d^\pi_t(S),\pi_t(A|S)}\left[  \|\nabla \log\pi_t(A|S)\|^2\right]\nonumber\\
    &\leq \frac{\sigma}{(1-\gamma)^4}, 
\end{align}
where the last inequality is from \ref{ass:smooth}. 


\begin{theorem}
Assume $\epsilon \leq \eta_t^2$ for all $k$, then 
\begin{align*}
    \min_{t\leq T}\min\{\mathbb{E} \|\nabla V_t\|_{2}, \mathbb{E} \|\nabla V_t\|_{2}^2 \} \leq O\left(\frac{1}{\sqrt{T}}\right).
\end{align*}
\end{theorem}

From Lemma \ref{lem:bounded variance} and our assumption on $\epsilon$, we have
\begin{align*}
    \eta_t (1-\epsilon)\min\{\mathbb{E} \|\nabla V_t\|_{2}, \mathbb{E} \|\nabla V_t\|_{2}^2 \} \leq \mathbb{E} \left[ V_{t+1}\right] -  \mathbb{E} \left[ V_{t}\right] + \eta_t^2 C, 
\end{align*}
where $C>0$ is a constant.

Following a similar vanilla SGD analysis \cite{dutta2020discrepancy, dutta2023demystifying}, we obtain the results. 




\section{Experimental Settings \& Additional Results}\label{appendix:extra}
This section complements our empirical results in \S\ref{sec:experiments}. We start with highlighting the implementation details. 

\subsection{B.1 \;Hardware and Runtime Details}\label{appendix:impl}
All profiling experiments were executed on a server equipped with NVIDIA Tesla V100 GPUs. Wall‐clock times per 100K timesteps varied between 10–30 minutes, depending on the environment and base algorithm.  On average, each lookback/mixup/three‐points variant incurred an overhead of $\sim$20\% compared to vanilla training, given that we consider a minimum of five episodes for the evaluations for the results, resulting in superior training performance in most cases. As we note, increasing $E$ inflates the compute cost without proportional gains in stability or final performance. This sheds light on the practical trade-off that a modest number of rollouts suffices to stabilize learning while preserving wall-clock efficiency.

\subsection{B.2 \; Profiling Pseudocode}
Algorithm~\ref{alg:profiling} presents the reward‐profiling procedure used in all experiments. The framework wraps around any base policy-gradient update $\mathcal{U}$ and augments it with one of three selection strategies: lookback (LB), mixup (MU), or three-points (TP). Each variant compares candidate policies using the empirical return $\hat{J}_E$ computed from $E$ i.i.d.\ evaluation rollouts, and retains the highest‐performing candidate according to its selection rule.

\begin{algorithm}[ht]
\caption{Reward Profiling Framework used in implementation} 
\label{alg:profiling}
\SetKwInOut{Input}{Input}\SetKwInOut{Output}{Output}
\Input{initial $\theta_0$; iterations $T$; rollouts $E$; mix weight $\lambda$ (or $\lambda\!\sim\!\mathrm{Beta}(\alpha,\beta)$); base update $\mathcal U$; variant $\in\{\mathrm{LB},\mathrm{MU},\mathrm{TP}\}$}
\Output{$\theta_T$}

Let $\hat J_E(\phi) \;=\; \frac{1}{E}\sum_{i=1}^{E} G(\tau_i), \quad \tau_i \sim \pi_\phi$

\For{$t=0$ \KwTo $T-1$}{
  $\theta' \leftarrow \mathcal U(\theta_t)$\;
  \If{variant $\neq$ LB}{ $\theta^{\mathrm{mix}} \leftarrow \lambda\,\theta' + (1-\lambda)\,\theta_t$ }
  \uIf{variant = LB}{ $\mathcal C \leftarrow \{\theta_t,\theta'\}$ }
  \uElseIf{variant = MU}{ $\mathcal C \leftarrow \{\theta_t,\theta^{\mathrm{mix}}\}$ }
  \Else{ $\mathcal C \leftarrow \{\theta_t,\theta',\theta^{\mathrm{mix}}\}$ }
  $\theta_{t+1} \leftarrow \arg\max_{\phi\in\mathcal C}\ \hat J_E(\phi)$ with tie-break to $\theta_t$\;
}
\Return{$\theta_T$}
\end{algorithm}

\subsection{B.3 \; Environments}
\subsubsection{Gymnasium and Bullet}
\label{appendix:mujoco-bullet}

We evaluate our methods on a suite of standard continuous control tasks with PyBullet environments, including Ant, Humanoid, Hopper, HalfCheetah, and Walker2D. These feature high-dimensional state and action spaces, complex multi-body dynamics, and challenging reward structures, while the Box2D suite provides 2D physics tasks with moderate state and action dimensionality.

All experiments in these environments use default hyperparameters from Stable Baselines3 unless mentioned otherwise, summarized in Table~\ref{tab:hyperparams} to ensure fair comparison and reproducibility across all algorithms and tasks.

\begin{table}
\centering
\caption{Default Hyperparameters for MuJoCo/Bullet Experiments}
\label{tab:hyperparams}
\begin{tabular}{llll}
\toprule
\textbf{Parameter} & \textbf{PPO} & \textbf{DDPG} & \textbf{TRPO} \\
\midrule
Learning Rate & $3\times10^{-4}$ & $1\times10^{-3}$ & $1\times10^{-3}$ \\
Network Architecture & [64,64] MLP & [400,300] MLP & [64,64] MLP \\
Batch Size & 64 & 100 & 128 \\
Discount ($\gamma$) & 0.99 & 0.99 & 0.99 \\
\bottomrule
\end{tabular}
\end{table}

\begin{figure*}
  \centering
  \includegraphics[width=0.23\linewidth]{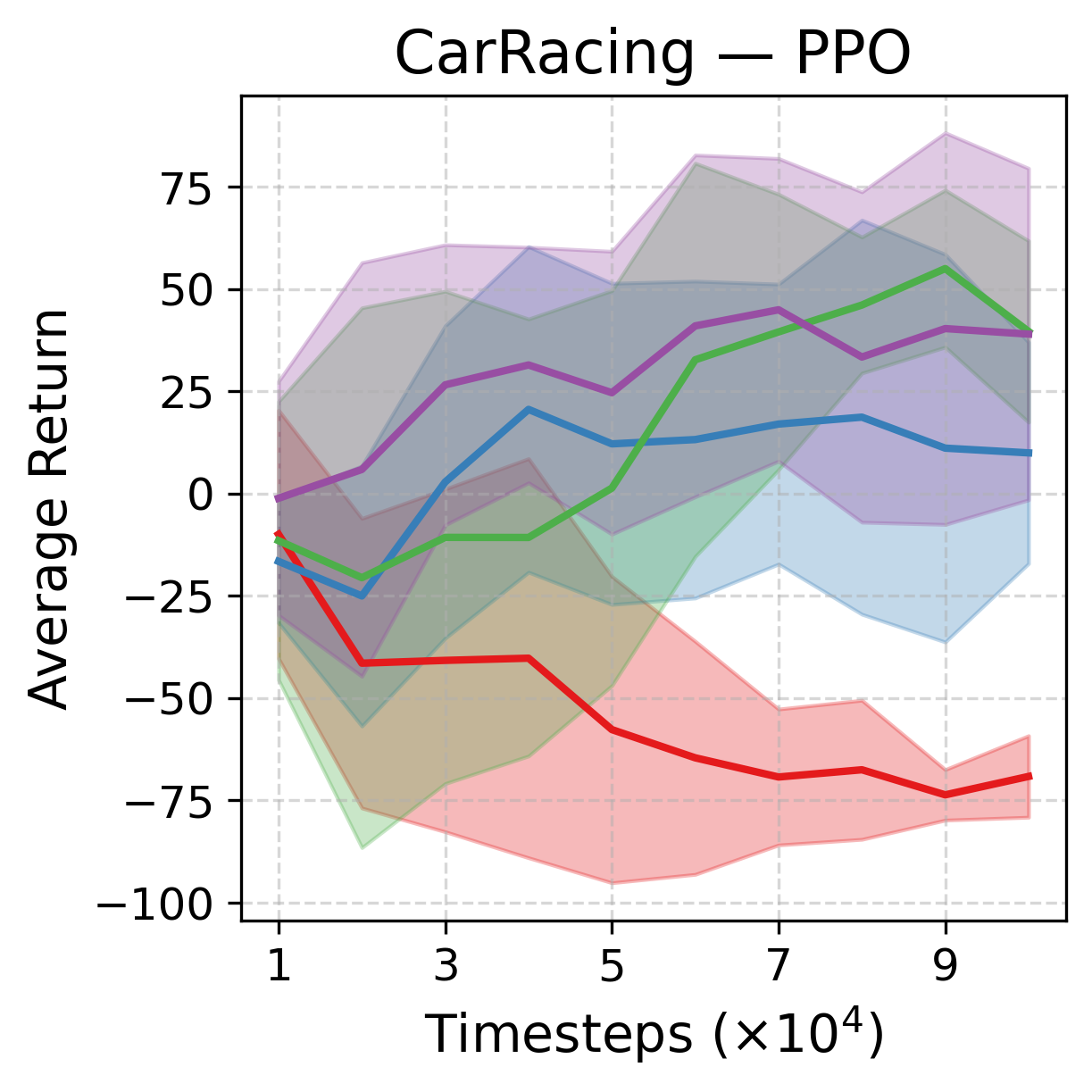}\hfill
  \includegraphics[width=0.23\linewidth]{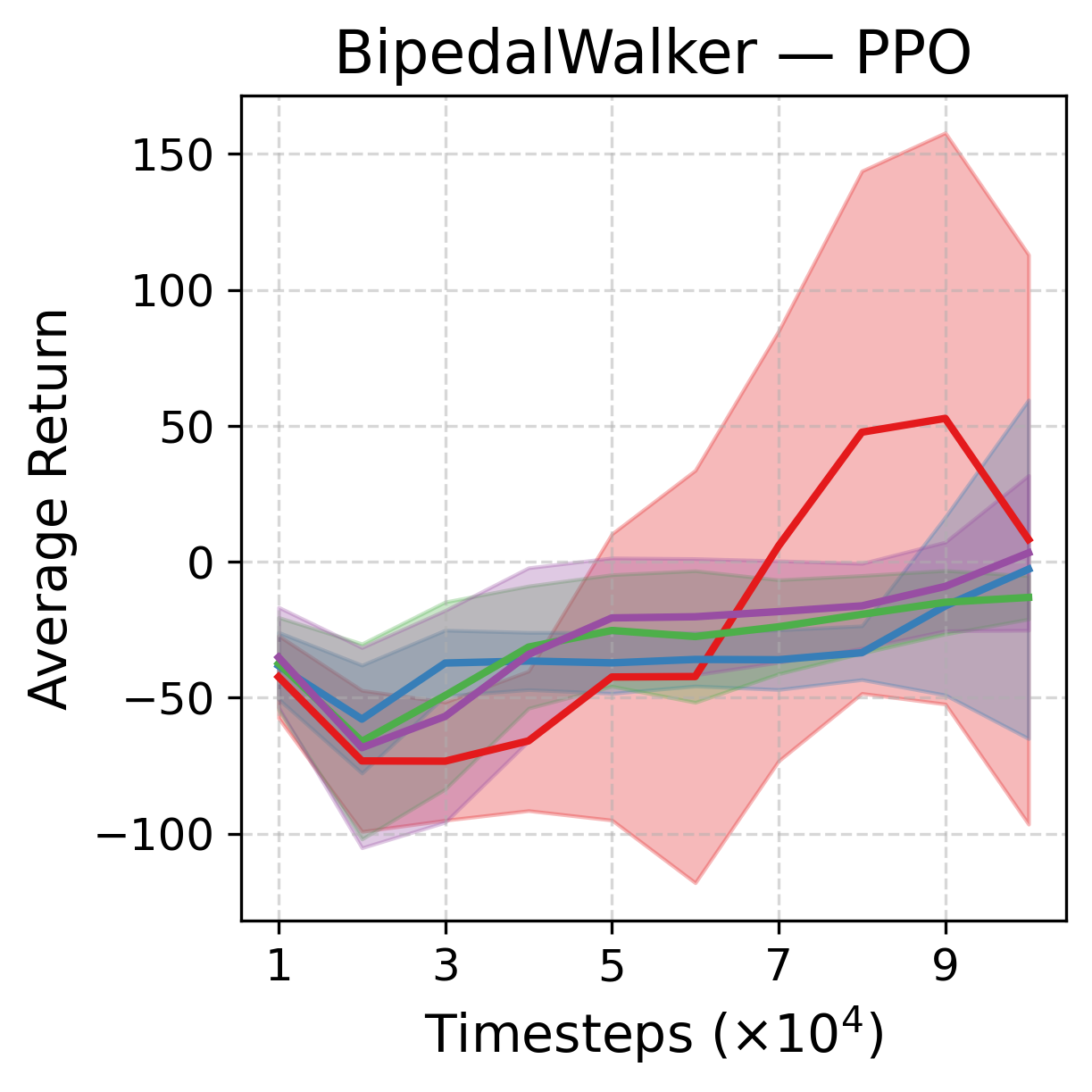}\hfill
  \includegraphics[width=0.23\linewidth]{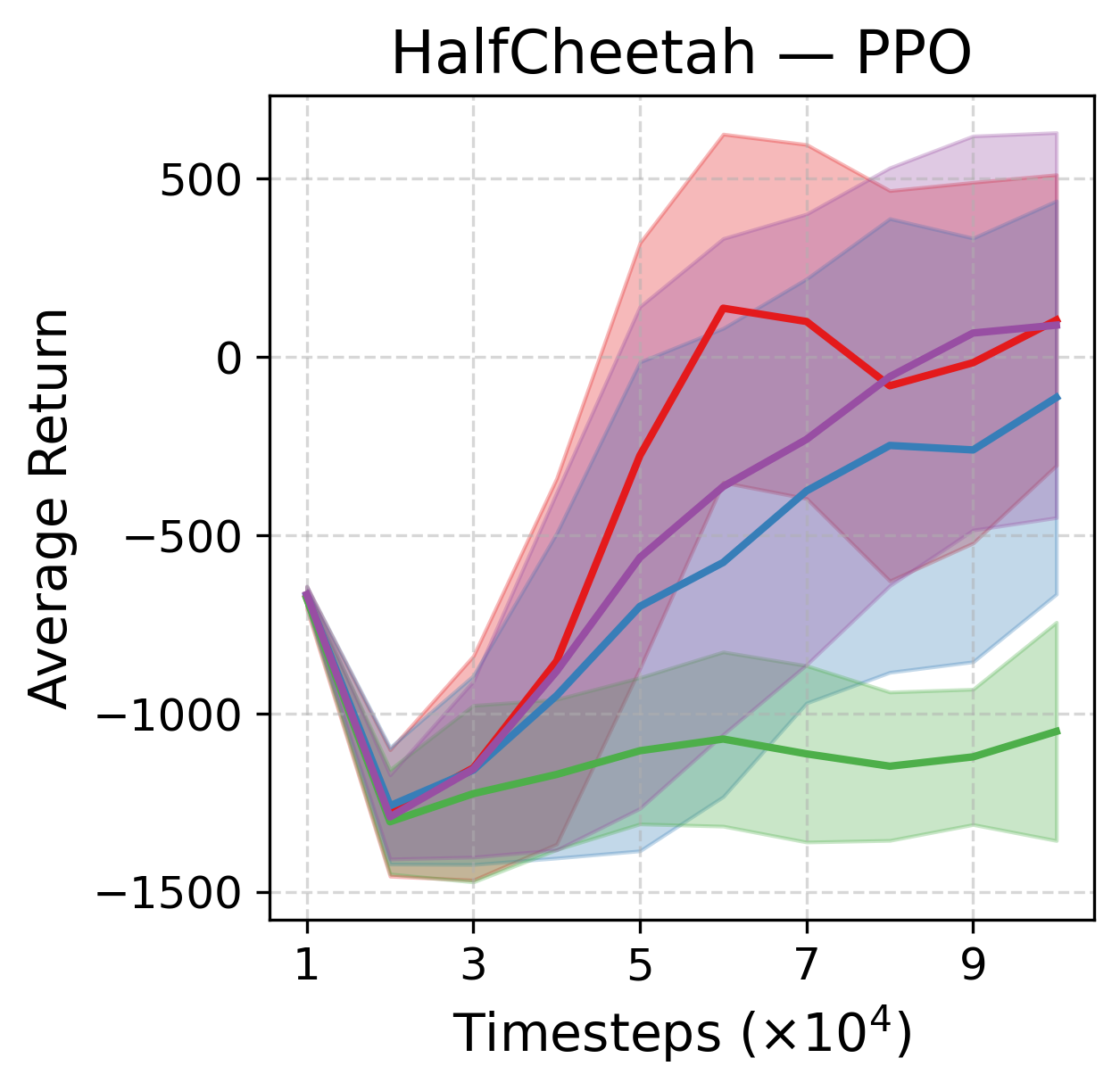}\hfill
  \includegraphics[width=0.23\linewidth]{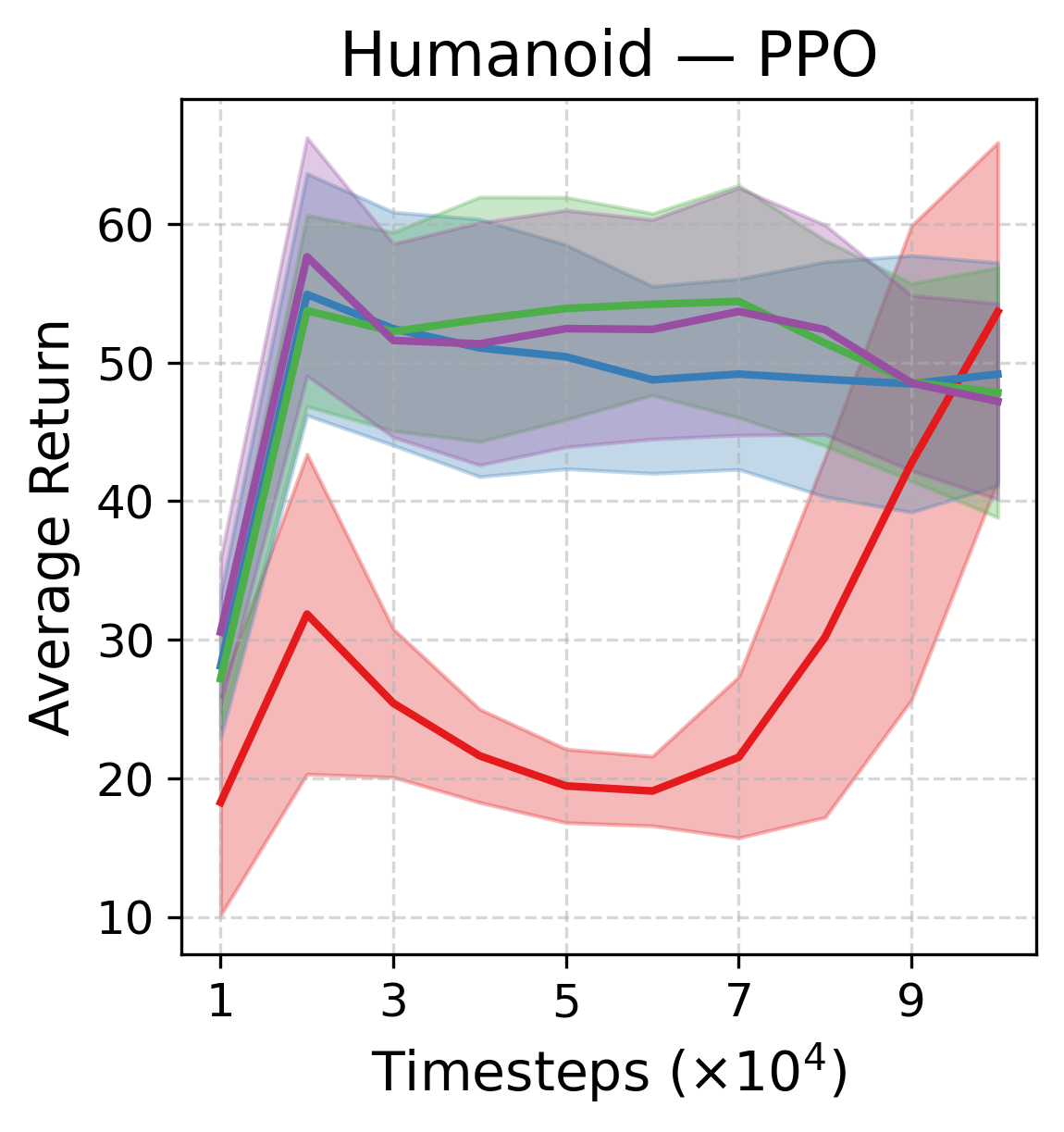}

  \medskip
  \includegraphics[width=0.23\linewidth]{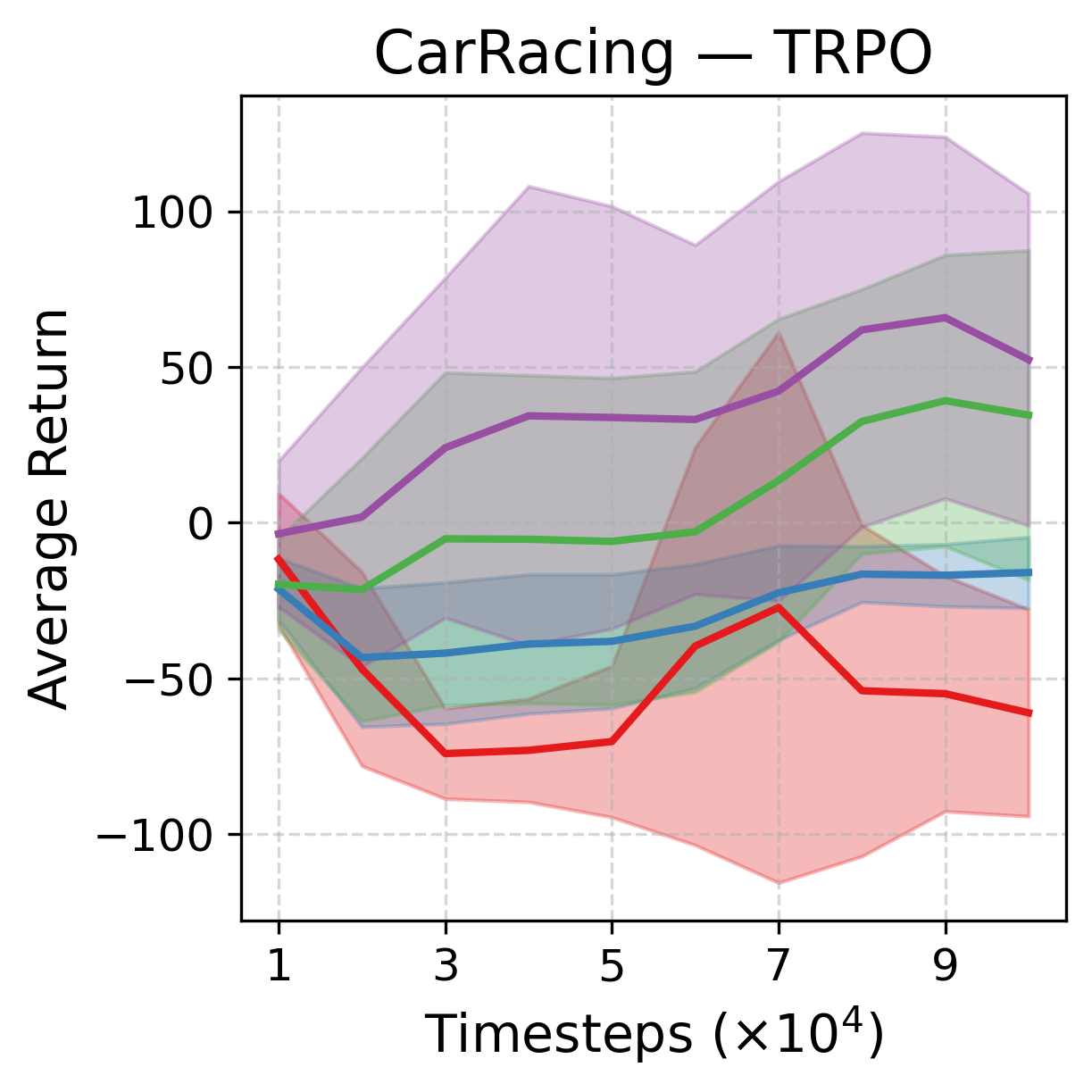}\hfill
  \includegraphics[width=0.23\linewidth]{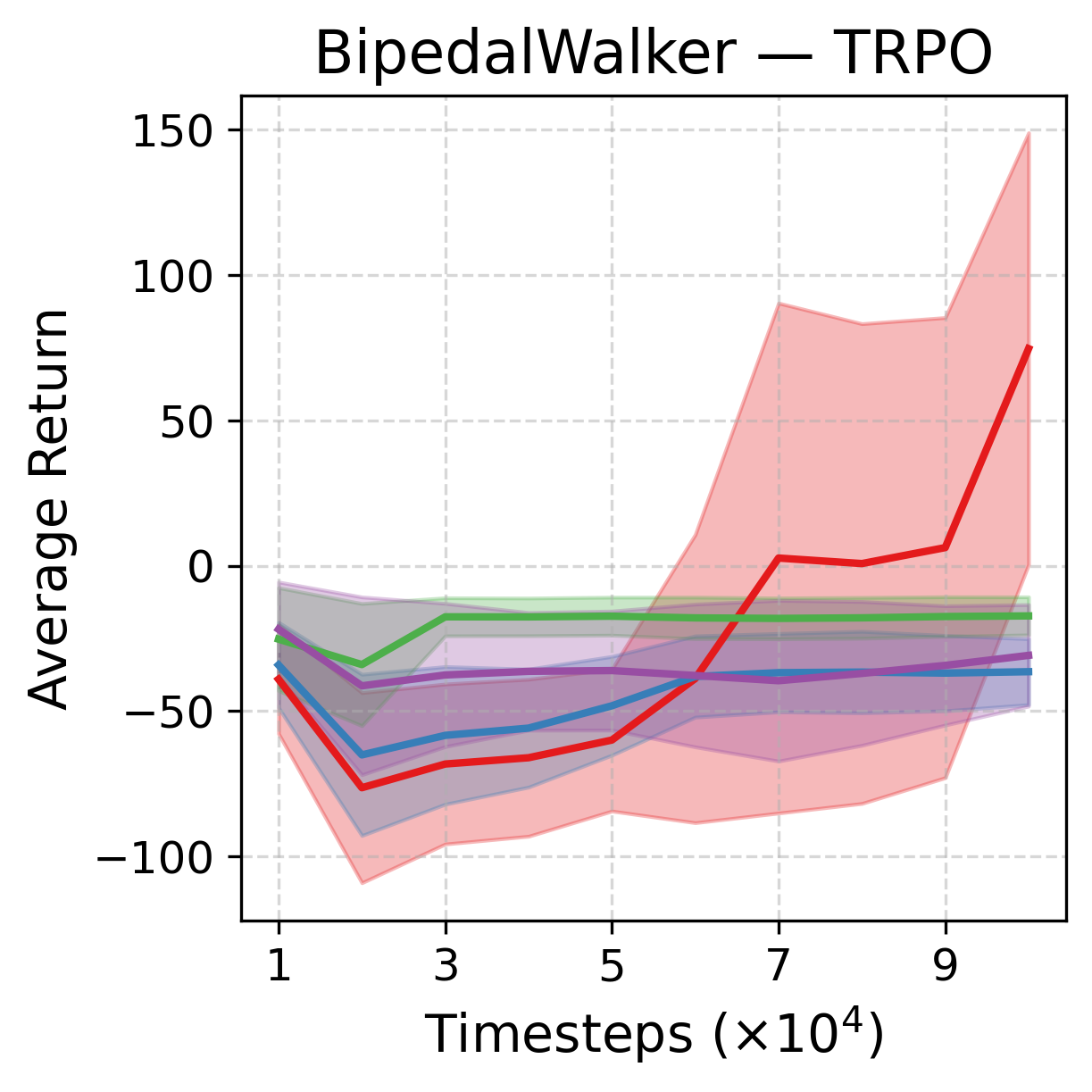}\hfill
  \includegraphics[width=0.23\linewidth]{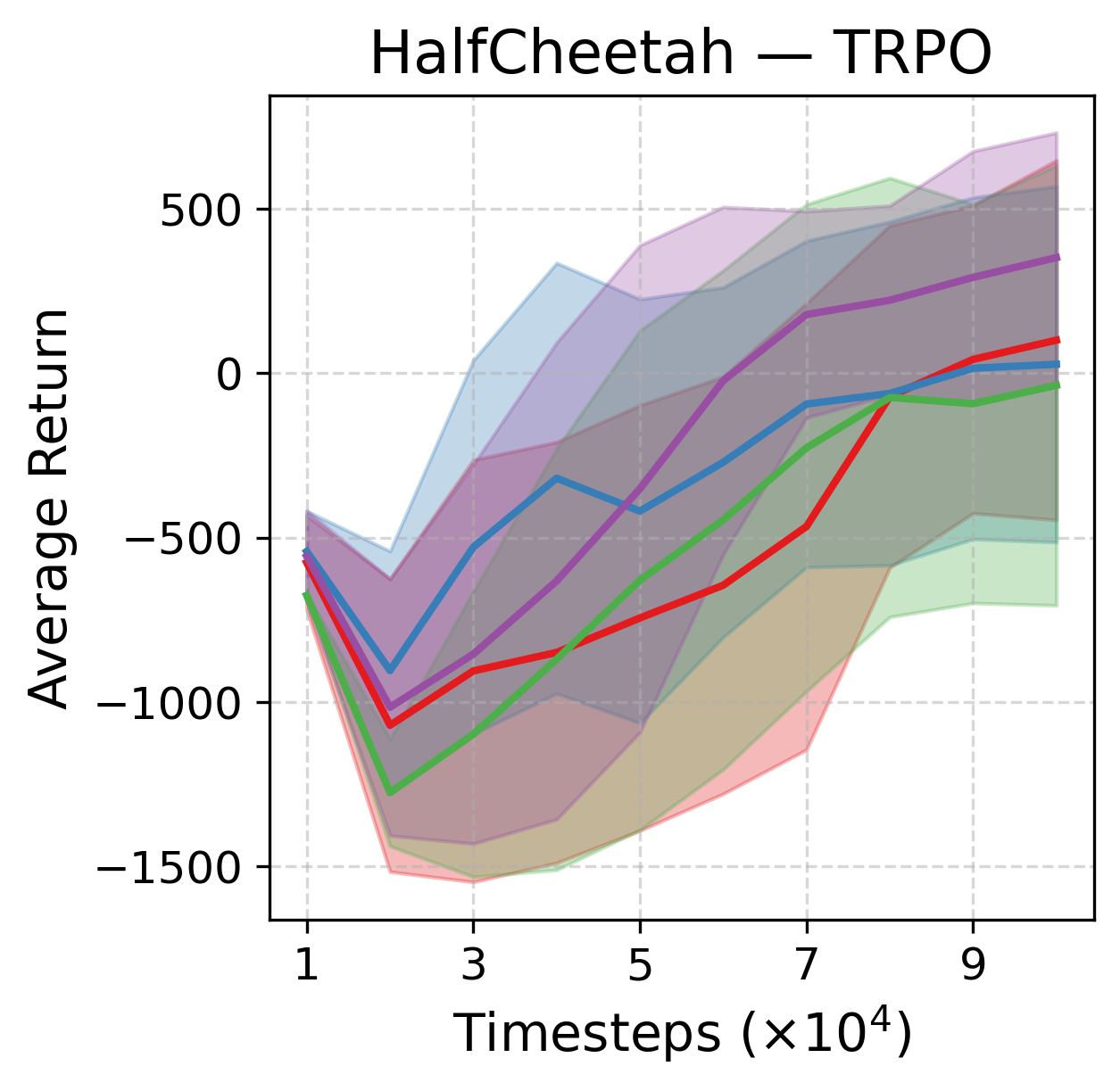}\hfill
  \includegraphics[width=0.23\linewidth]{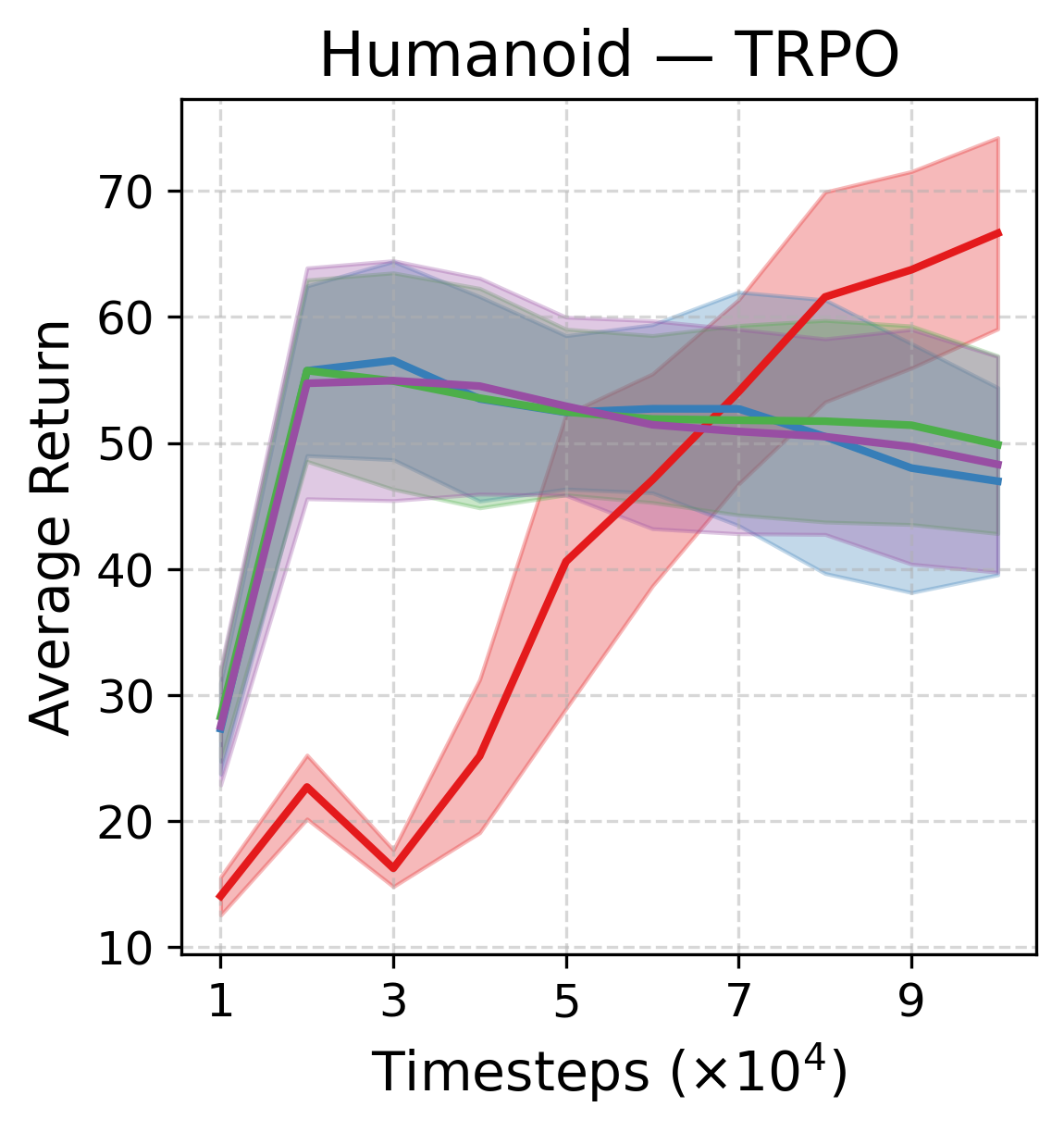}

  \medskip
  \includegraphics[width=0.23\linewidth]{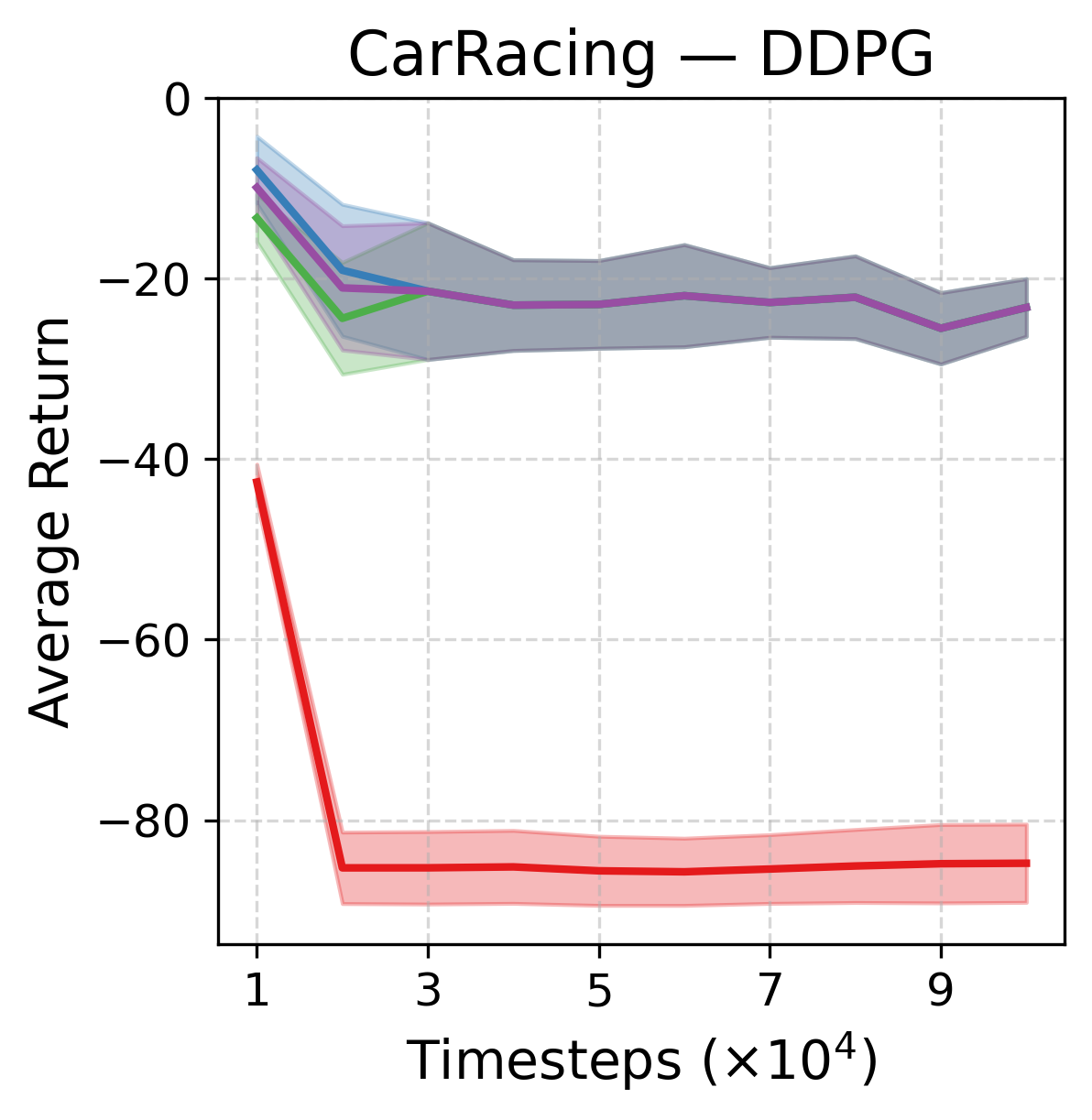}\hfill
  \includegraphics[width=0.23\linewidth]{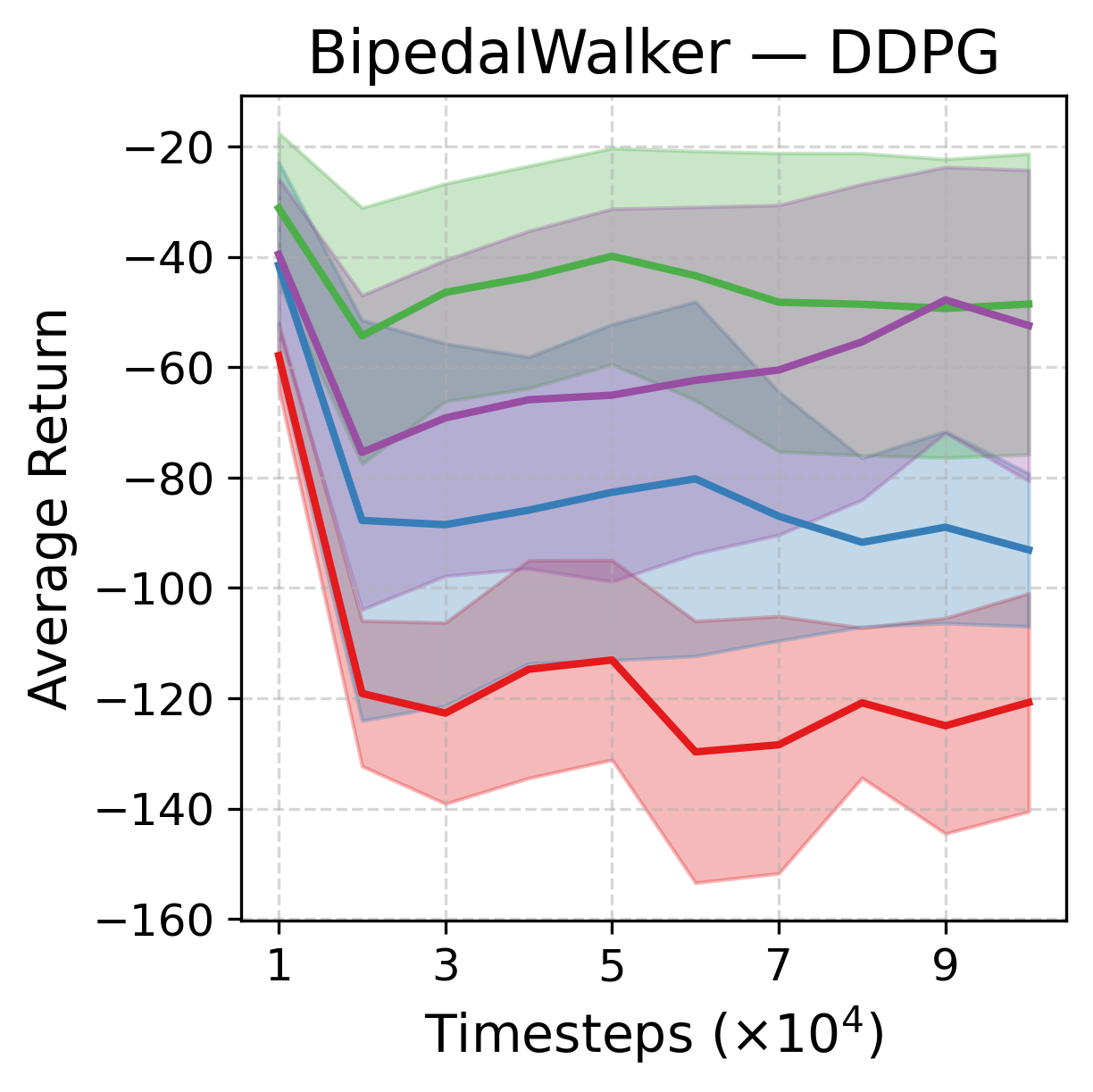}\hfill
  \includegraphics[width=0.23\linewidth]{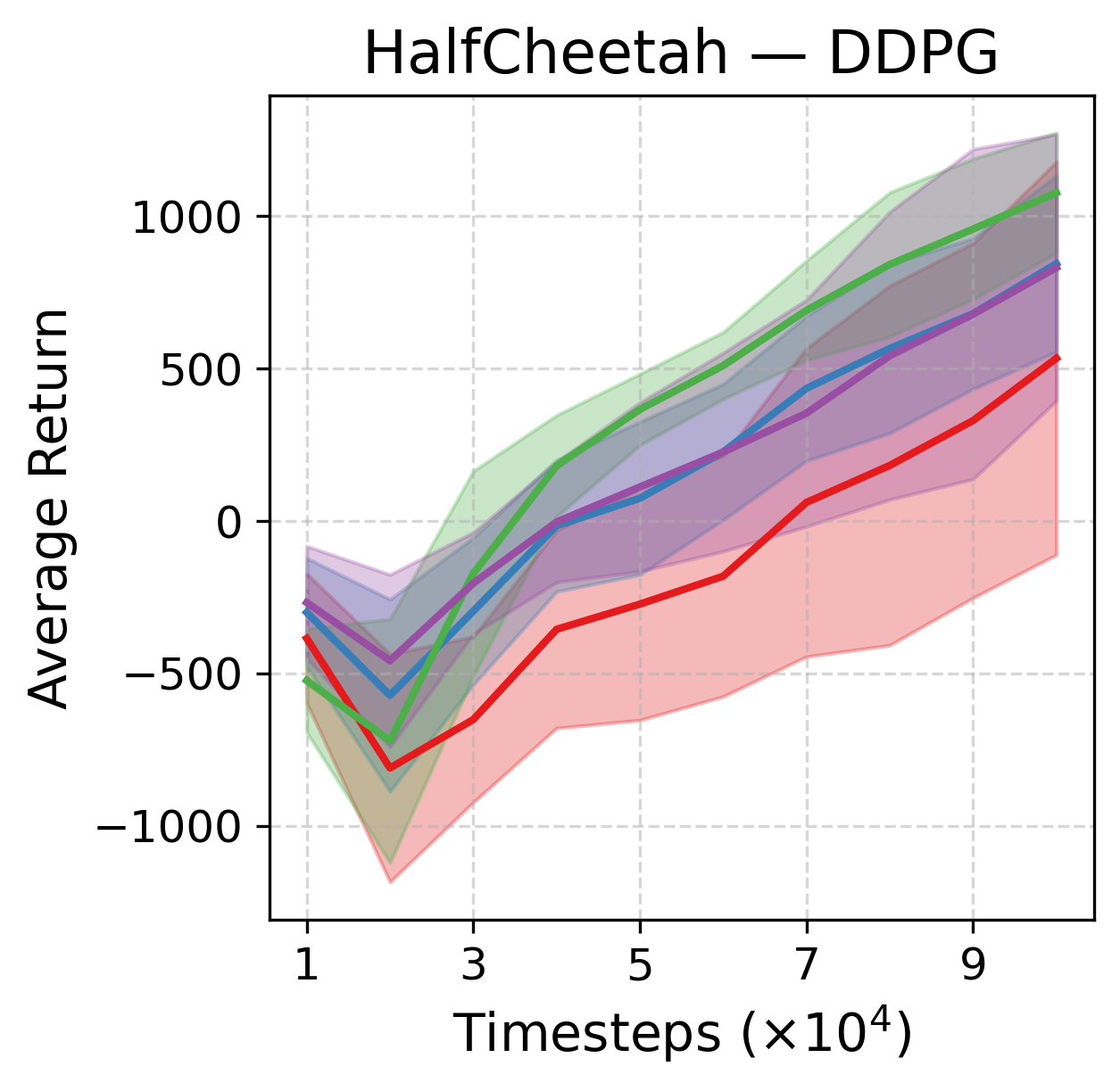}\hfill
  \includegraphics[width=0.23\linewidth]{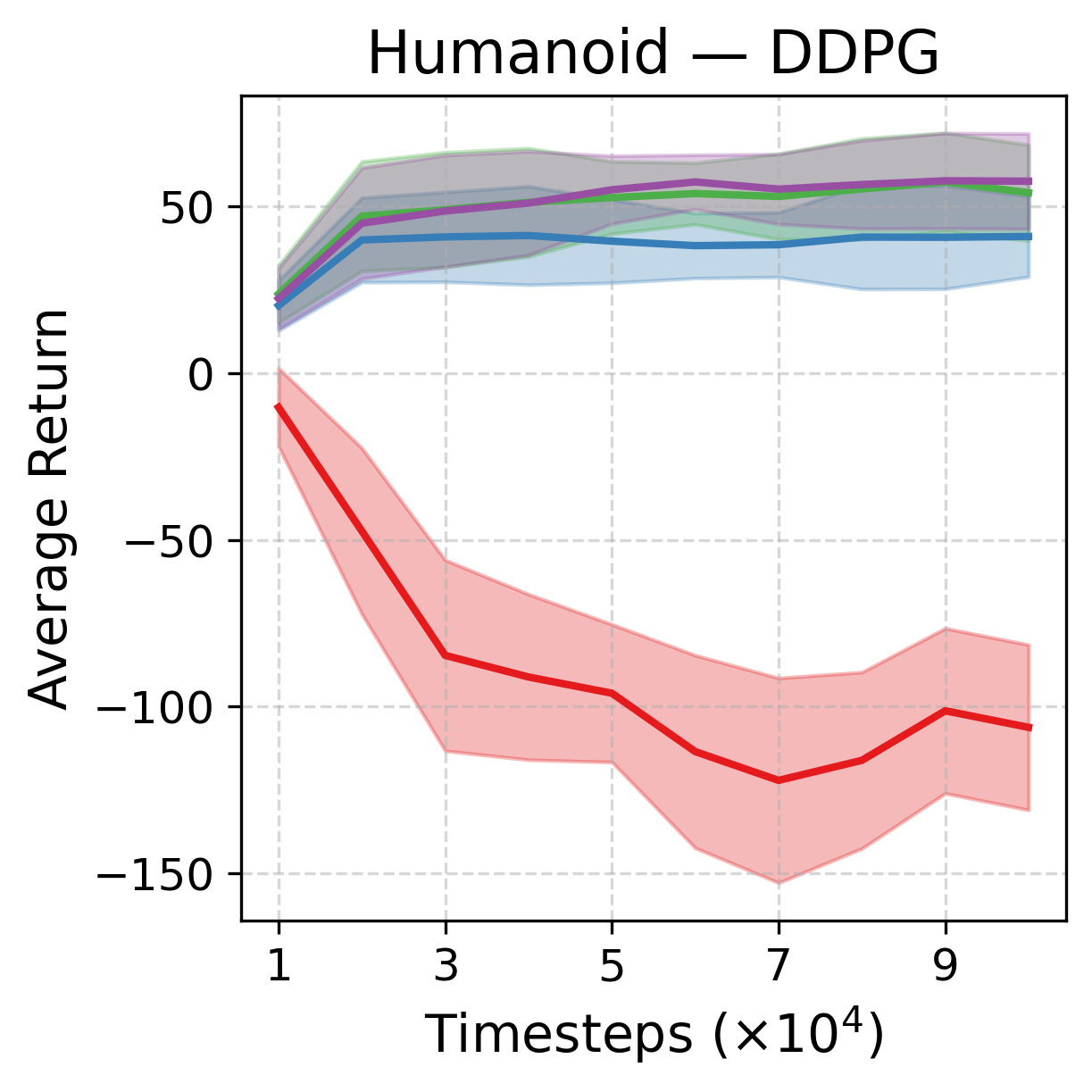}

  \medskip
  \hspace*{0.13\linewidth}
  \includegraphics[width=0.23\linewidth]{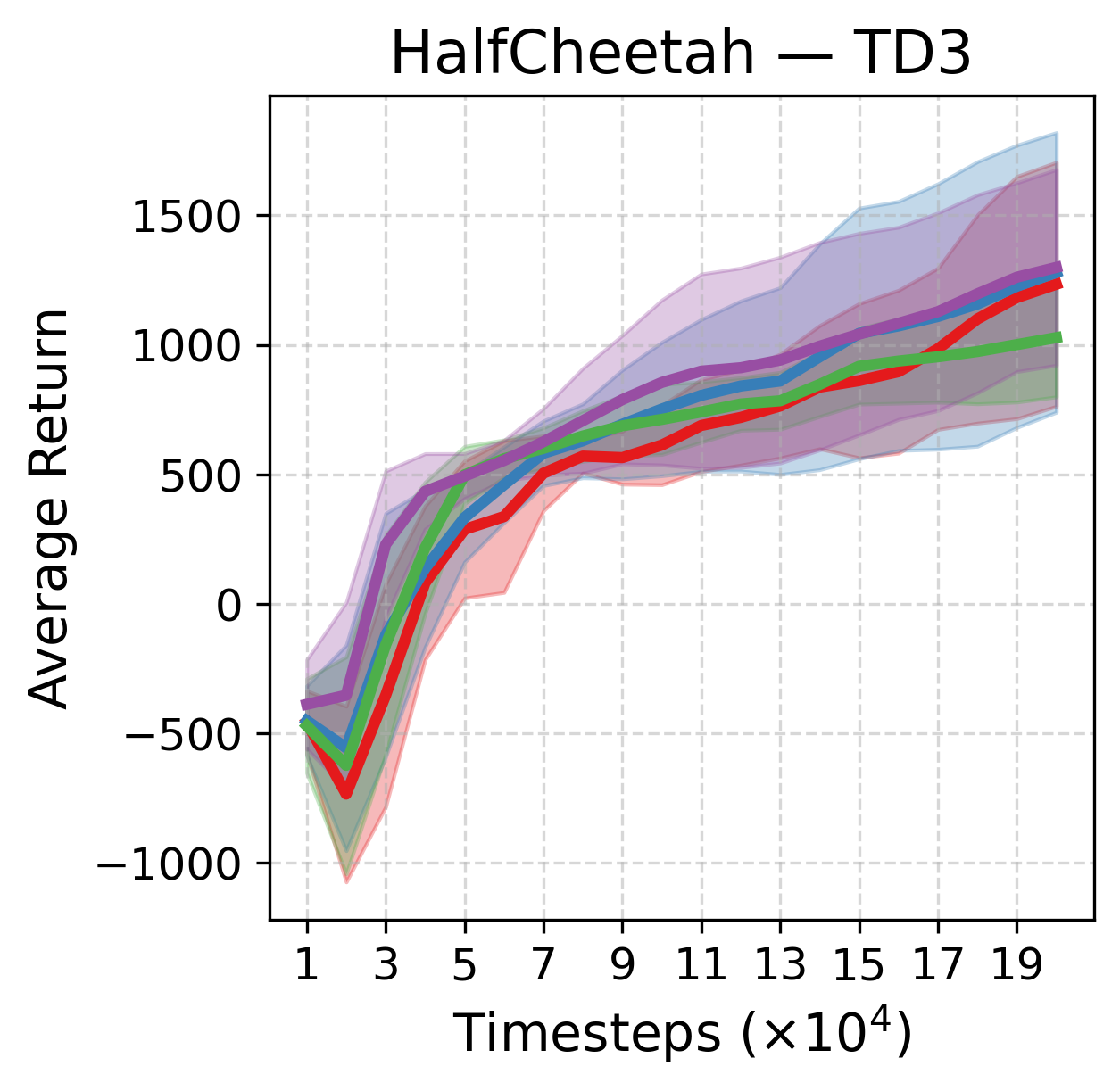}\hfill
  \includegraphics[width=0.23\linewidth]{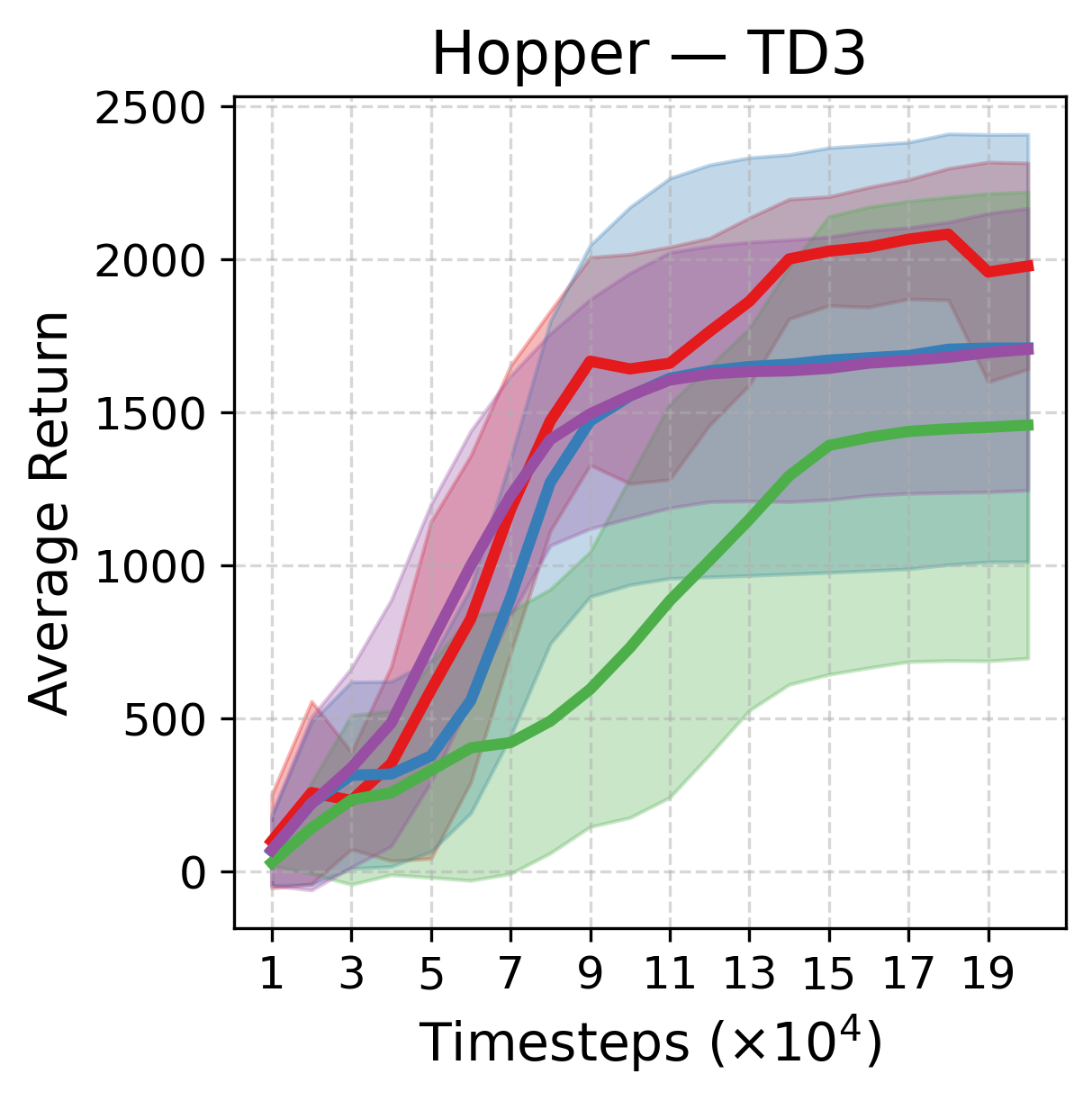}\hfill
  \includegraphics[width=0.23\linewidth]{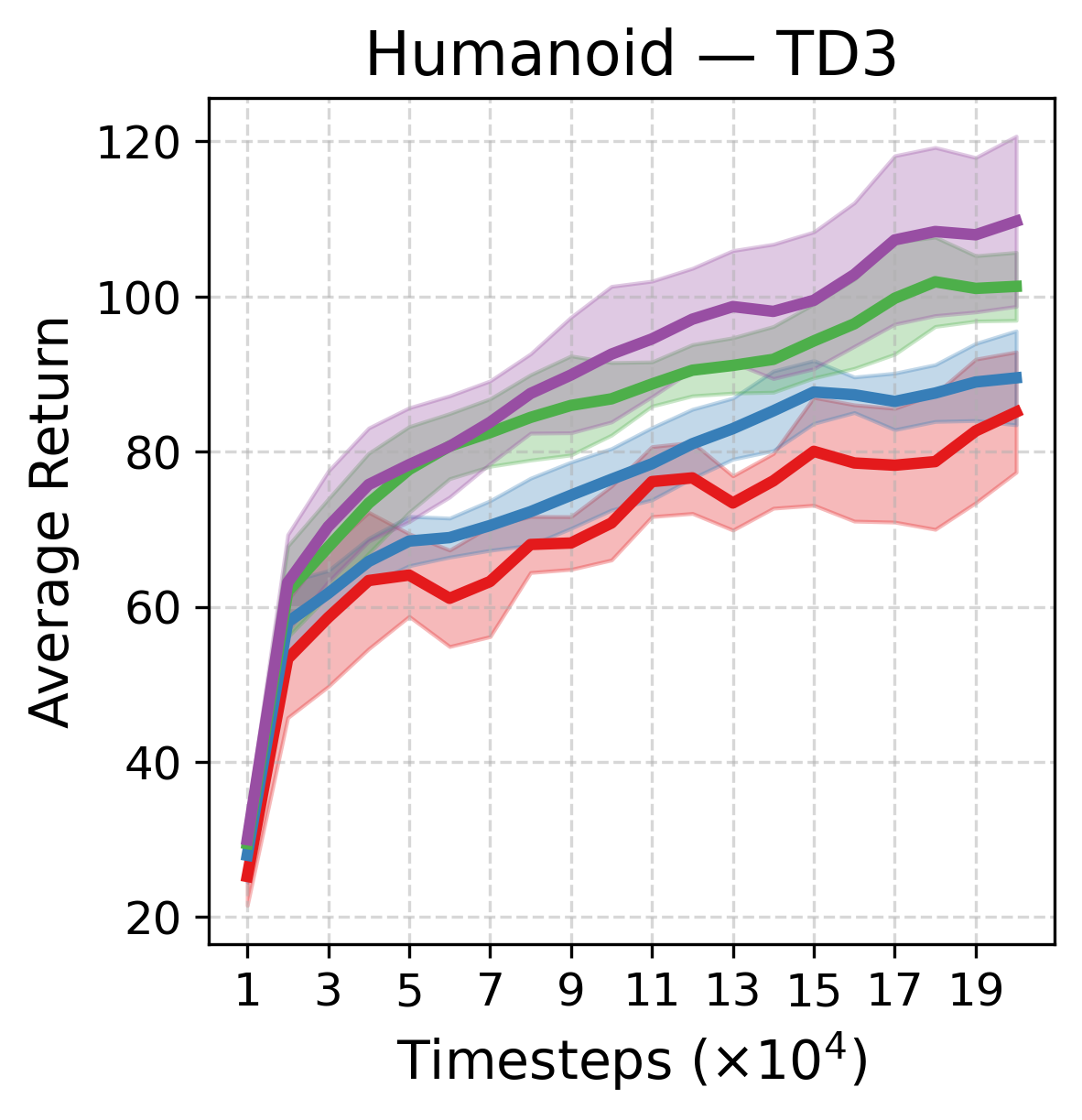}
  \hspace*{0.13\linewidth}

  \caption{
    Average return vs.\ profiling iteration for PPO (top), TRPO (middle), DDPG (3rd row),
    across CarRacing, BipedalWalker, HalfCheetah, and Humanoid.
    The bottom row shows TD3 performance on three Bullet‐locomotion tasks.
     variants: 
    \textcolor[HTML]{e41a1c}{\rule[0.5ex]{1.5em}{1pt}} Vanilla, 
    \textcolor[HTML]{377eb8}{\rule[0.5ex]{1.5em}{1pt}} Lookback, 
    \textcolor[HTML]{4daf4a}{\rule[0.5ex]{1.5em}{1pt}} Mixup, 
    \textcolor[HTML]{984ea3}{\rule[0.5ex]{1.5em}{1pt}} Three‐Points}\vspace{-2em}
  \label{fig:appendix_additional_curves}
\end{figure*}

\subsubsection{Unity ML-Agents Reacher}
\label{appendix:unity}

Here, \emph{20} independent robotic arms operate in parallel, each receiving a 33-dimensional observation vector (joint angles, joint velocities, target position) and producing 4-dimensional continuous torques.  At each timestep, an agent earns a dense reward of +0.1 for keeping its fingertip within the moving target zone, encouraging precise coordination of exploration and control.

We wrap our Three-Points (TP) profiling layer around a decentralized DDPG backbone.  Every \emph{50} episodes, we evaluate each agent’s policy on $E=5$ rollouts and perform consensus voting: the new, old, and mixed candidates are compared across the 5 agents, and the majority vote selects the next shared policy parameters to strike balances in timely policy correction with scalable multi-agent evaluation. The environment's deterministic physics and reward structure allow DDPG to achieve target performance in fewer than 110 episodes, significantly faster than in MuJoCo or Bullet tasks.

\subsection{B.4 \; Additional Experiments}

In the main paper (Figure~\ref{fig:all_return_curves}), we showed four continuous‐control suites for PPO, TRPO, and DDPG.  Here, we complete the picture by adding results on two Box2D tasks (CarRacing, BipedalWalker) and two high‐DOF Bullet tasks (HalfCheetah, Humanoid). We also evaluated our profiling wrapper on the state-of-the-art off-policy actor-critic, TD3 \citep{fujimoto18addressing}, that extends DDPG with clipped-double Q-learning, delayed policy updates, and target-policy smoothing.  All of TD3’s core improvements—reduced overestimation bias from twin critics, more accurate actor gradients via delayed updates, and smoother targets via policy noise—are preserved inside our wrapper while reward profiling yields similar stability gains across both on‐ and off‐policy methods.

\subsubsection{Extension to Off‐Policy Actor–Critics}

\begin{figure}[H]
  \centering
  \includegraphics[width=0.65\linewidth]{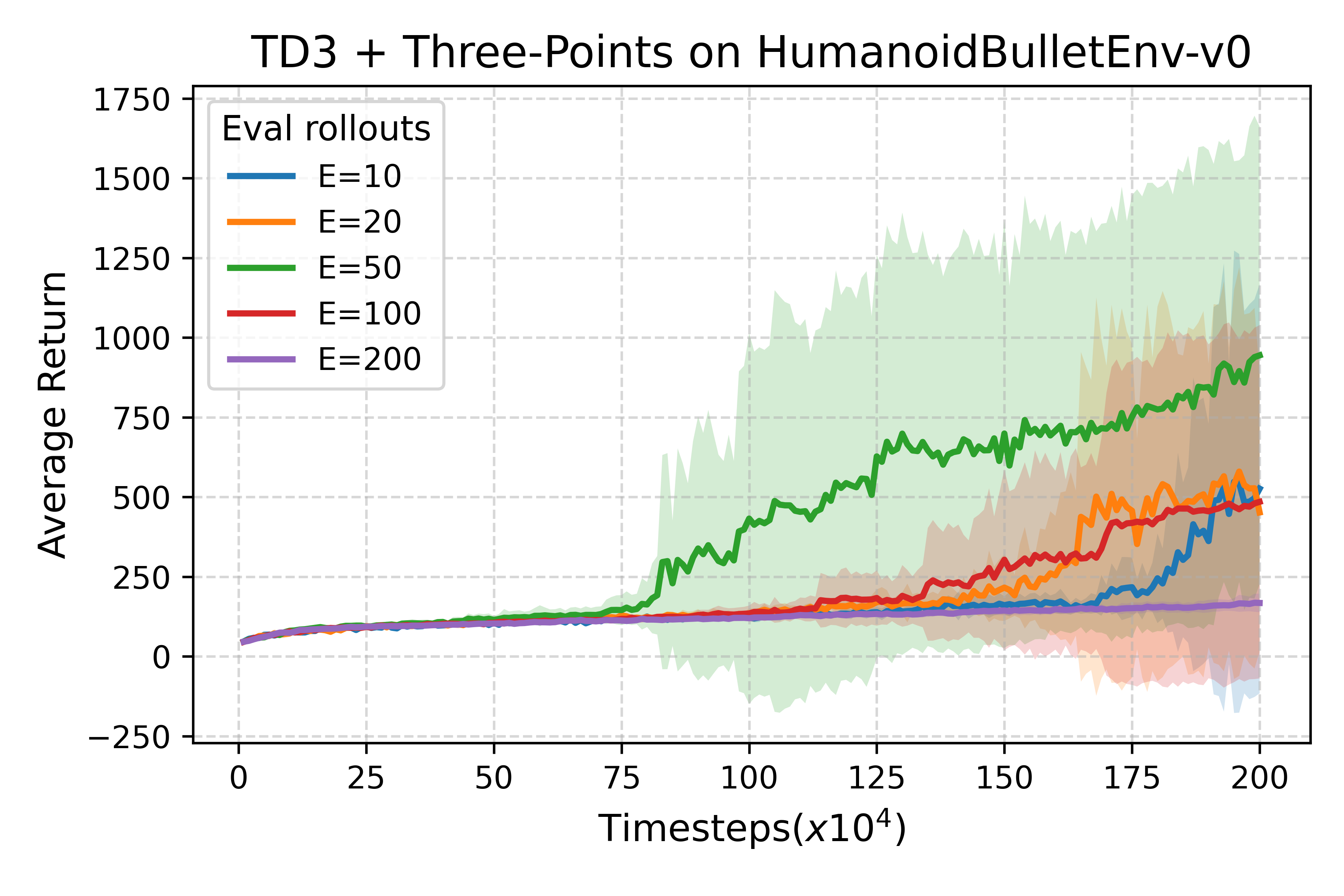}
  \caption{\small
    \textbf{TD3 + Three-Points on Humanoid.}
    Training performance for TD3 wrapped with our Three-Points variant, as the number of eval rolls \(E\) varies.  
    \(\mathbf{E=10,20,50,100,200}\) are shown in distinct colors.  
    Small \(E\) (10,20) yields erratic updates; large \(E\) (200) stabilizes but delays progress; intermediate \(E\!\approx\!50\)–100 offers the best trade-off.
  }
  \label{fig:td3_humanoid}
\end{figure}

While our main evaluations focused on PPO, TRPO, and DDPG across multiple benchmarks, the reward-profiling wrapper is agnostic to the inner loop.  To illustrate, we slot TD3 into our framework and test it on \textbf{HumanoidBulletEnv-v0}. No changes to the outer logic are required: the TD3 “Step” call simply replaces DDPG’s. Figure~\ref{fig:td3_humanoid} repeats the sensitivity analysis for TD3+3P on HumanoidBulletEnv-v0.  We observe the same “U‐shaped’’ trade‐off in evaluation budget \(E\). Comparatively smaller evaluation budgets (\(E=10,20\)) yield highly noisy reward estimates, leading to erratic policy corrections and sluggish long-term improvement. At the other extreme, very large budgets (\(E=200\)) produce overly conservative updates: the low variance rarely permits the policy to depart from its initialization, resulting in nearly flat learning curves. Intermediate budgets (\(E=50\)–100) achieve the best of both worlds, taming erratic decision-making while still allowing sufficient exploration—\(E=50\) reaches the highest peak returns, and \(E=100\) delivers a steadier ascent at only marginally lower peak performance. This empirical pattern echoes our high-probability monotonicity analysis (Appendix~\ref{appendix:mono-hp}): increasing \(E\) shrinks the confidence width on the estimated return at rate \(\mathcal{O}(e^{-2E\epsilon^2/B^2})\), but beyond a critical threshold this conservatism prevents the sampler from accepting bold, potentially beneficial updates. In practice, we therefore recommend moderate evaluation budgets to balance stability and learning speed.

\begin{table}
\centering
\caption{DDPG Hyperparameters for Unity Reacher}
\label{tab:unity-hparams}
\begin{tabular}{ll}
\toprule
\textbf{Parameter} & \textbf{Value} \\
\midrule
Learning Rate & $1\times10^{-3}$ \\
Batch Size & 128 \\
Replay Buffer Size & $1\times10^6$ \\
Tau ($\tau$) & 0.005 \\
Gamma ($\gamma$) & 0.99 \\
Actor Network & [400, 300] MLP \\
Critic Network & [400, 300] MLP \\
OU Noise ($\sigma$) & 0.2 \\
\bottomrule
\end{tabular}
\end{table}

\end{document}